\documentclass[11pt,a4paper]{article} 
\usepackage[margin=2.2cm]{geometry}

\usepackage[resetlabels]{multibib}
\newcites{app}{References}

\usepackage[utf8]{inputenc} 
\usepackage[T1]{fontenc}    
\usepackage{hyperref}
\usepackage{url}            
\usepackage{booktabs}       
\usepackage{amsfonts}       
\usepackage{nicefrac}       
\usepackage{microtype}      
\usepackage{xcolor}         

\definecolor{dgreen}{RGB}{65, 171, 93}

\hypersetup{
    colorlinks=true,
    linkcolor=blue,
    filecolor=magenta,      
    urlcolor=cyan,
    citecolor=dgreen,
    }     

\usepackage{bookmark}
\usepackage{mathtools}
\usepackage{mathalfa}
\usepackage{mathrsfs}
\usepackage{bm,bbm}
\usepackage{ulem}
\usepackage{textcomp}
\usepackage{gensymb}
\usepackage{floatrow}
\usepackage{enumerate}
\usepackage{algorithmic}
\usepackage{algorithm}
\usepackage{soul}
\usepackage{tikz}
\usepackage{tkz-euclide}
\usepackage{appendix}

\usepackage{chngcntr}
\counterwithin{figure}{section}

\usepackage{amssymb,amsmath,amsthm}
\usepackage{thmtools, thm-restate}

\numberwithin{equation}{section}

\input m.macros
\input m.symbols

\newtheorem{lemma}{Lemma}[section]

\usepackage{natbib}
\setcitestyle{round}

\usepackage[capitalize]{cleveref}

\newcommand{\DKL}[2]{\kl\left [#1,#2 \right]}

\title{Minimum Description Length Control}

\author{Ted Moskovitz\textsuperscript{1,*}, Ta-Chu Kao\textsuperscript{1,3}, Maneesh Sahani$^{1,\dagger}$, Matthew M. Botvinick$^{1,2,\dagger}$ 
\\
\\
{\normalsize 1. Gatsby  Unit, UCL} \\
 {\normalsize 2. DeepMind} \\
 {\normalsize 3. Facebook Reality Labs} \\
 {\normalsize $\dagger$ Co-senior authors.}\\
{\normalsize *Correspondence: \texttt{ted@gatsby.ucl.ac.uk}}
}
\date{}

\begin{document}
\maketitle

\begin{abstract}
  We propose a novel framework for multitask reinforcement learning based on the \textit{minimum description length} (MDL) principle. In this approach, which we term \textit{MDL-control} (MDL-C), the agent learns the common structure among the tasks with which it is faced and then distills it into a simpler representation which facilitates faster convergence and generalization to new tasks. In doing so, MDL-C naturally balances adaptation to each task with epistemic uncertainty about the task distribution. We motivate MDL-C via formal connections between the MDL principle and Bayesian inference, derive theoretical performance guarantees, and demonstrate MDL-C's empirical effectiveness on both discrete and high-dimensional continuous control tasks. 
\end{abstract}

\section{Introduction}
In order to learn efficiently in a complex world with multiple,  sometimes rapidly changing objectives, both animals and machines must leverage information obtained from past experience. This is a challenging task, as processing and storing all relevant information is computationally infeasible. How can an intelligent agent address this problem? We hypothesize that one route may lie in the \textit{dual process theory} of cognition, a longstanding framework in cognitive psychology first introduced by William James \citep{james1890principles} which lies at the heart of many dichotomies in both cognitive science and machine learning. Examples include goal-directed versus habitual behavior \citep{Graybiel:2008}, model-based versus model-free reinforcement learning \citep{daw:2011_twostep,Sutton:1998}, and ``System 1'' versus ``System 2'' thinking \citep{kahneman2011thinking}. In each of these paradigms, a complex, ``control'' process trades off with a simple, ``default'' process to guide actions. Why has this been such a successful and enduring conceptual motif? Our hypothesis is that default processes often serve to distill common structure from the tasks consistently faced by animals and agents, facilitating generalization and rapid learning on new objectives. For example, drivers can automatically traverse commonly traveled roads en route to new destinations, and chefs quickly learn new dishes on the back of well-honed fundamental techniques. Importantly, even intricate tasks can become automatic, if repeated often enough (e.g., the combination of fine motor commands required to swing a tennis racket): the default process must be sufficiently expressive to learn common behaviors, regardless of their complexity. In reality, most processes likely lie on a continuum between simplicity and complexity.

In reinforcement learning (RL; \citealp{Sutton:1998}), the problem of improving sample efficiency on new tasks is crucial to the developement of general agents which can learn effectively in the real world \citep{botvinick2015stats,zhang:2021_rlgen}. Intriguingly, one family of approaches which have shown promise in this regard are \textit{regularized policy optimization} algorithms, in which a goal-specific control \textit{policy} is paired with a simple yet general default \textit{policy} to facilitate learning across multiple tasks \citep{Teh:2017_distral,Galashov:2019,Goyal:2020,Goyal:2019,Moskovitz:2021_rpotheory}. One difficulty in algorithm design, however, is how much or how little to constrain the default policy, and in what way. An overly simple default policy will fail to identify and exploit commonalities among tasks, while an overly complex model may overfit to a single task and fail to generalize. Most approaches manually specify an asymmetry between the control and default policies, such as hiding input information \citep{Galashov:2019} or constraining the model class \citep{Lai:2021_compression}. Ideally, we'd like an adaptive approach that can learn the appropriate degree of complexity via experience. 

The \textit{minimum description length principle} (MDL; \citealp{rissanen_78mdl}), which in general holds that one should prefer the simplest model that accurately fits the data, offers a guiding framework for algorithm design that does just that, enabling the default policy to optimally trade off between adapting to information from new tasks and maintaining simplicity. 
Inspired by dual process theory and the MDL principle, we propose  \textit{MDL-control} (MDL-C, pronounced ``middle-cee''), a principled RPO framework for multitask RL. In \cref{sect:rl_intro}, we formally introduce multitask RL and describe RPO approaches within this setting. In \cref{sec:mdl-principle}, we describe MDL and the variational coding framework, from which we extract MDL-C and derive its formal performance characteristics. In \cref{sect:experiments}, we demonstrate its empirical effectiveness in both discrete and continuous control settings. Finally, we discuss related ideas from the the literature (\cref{sect:related_work}) and conclude (\cref{sect:conclusion}). 

\section{Reinforcement Learning Preliminaries} \label{sect:rl_intro}
%
\paragraph{Notation} In the following, we use $\DKL{p}{q}$ to denote the Kullback-Leibler divergence from distributions $q$ to $p$.
We use $\mathcal{N}(x; \mu, \sigma^2)$ to denote a normal distribution with mean $\mu$ and variance $\sigma^2$ for variable $x$.
We use $\delta(x)$ to refer to the Dirac-delta function.
\paragraph{The single-task setting} We model a task as a \textit{Markov decision process} (MDP;~\citealp{Puterman:2010}) $M = (\St, \A, \pr, r, \gamma, \rho)$, where $\St,\A$ are state and action spaces, respectively, $\pr: \St \times \A \to \mathcal P(\St)$ is the state transition distribution, $r: \St \times \A \to [0,1]$ is a reward function, $\gamma\in[0,1)$ is a discount factor, and $\rho \in \mathcal P(\St)$ is the starting state distribution. $\mathcal P(\cdot)$ is the space of probability distributions defined over a given space. The agent takes actions using a \textit{policy} $\pi: \St \to \mathcal P(\A)$. In large or continuous domains, the policy is often parameterized: $\pi \to \pi_\theta, \ \theta\in\Theta$, where $\Theta\subseteq\reals^d$ represents a particular model class with $d$ parameters. In conjunction with the transition dynamics, the policy induces a distribution over trajectories $\tau = (s_h, a_h)_{h=0}^\infty$, $\pr^{\pi_\theta}(\tau)$.
In a single task, the agent seeks to maximize its \textit{value}  $V^{\pi_\theta} = \E_{\tau\sim\pr^{\pi_\theta}}R(\tau)$, where $R(\tau) \coloneqq \sum_{h\geq0} \gamma^h r(s_h, a_h)$ is called the \textit{return}. We denote by $d^\pi_\rho$ the state-occupancy distribution induced by policy $\pi$ with starting state distribution $\rho$: $d_\rho^\pi(s) = \E_{\rho} (1 - \gamma) \sum_{h\geq 0}\gamma^h \Pr(s_h=s|s_0)$. 

\paragraph{Multiple tasks} In standard multitask RL, there is a (possibly infinite) set of tasks (MDPs) $\mathcal M = \{M\}$, usually presented to the agent by sampling from some task distribution $\pr_\mathcal M \in \mathcal P(\mathcal M)$. 
Typical objectives include finding either a single policy or a set of policies which maximize worst- or average-case value: $\max_\pi \min_{M\in \mathcal M} V_M^\pi$ \citep{Zahavy:2021_smp} or $\max_\pi \E_{\pr_\mathcal M} V_M^\pi$ \citep{Moskovitz:2021_rpotheory}. When the emphasis is on decreasing the required sample complexity of learning new tasks, a useful metric is \textit{cumulative regret}: the agent's total shortfall across training compared to an optimal agent.  In practice, it's often simplest to consider the task distribution $\pr_\mathcal M$ to be a categorical distribution defined over a discrete set of tasks $\mathcal M \coloneqq \{M_k\}_{k=1}^K$, though continuous densities over MDPs are also possible. Two multitask settings which we consider here are \textit{parallel task} RL and \textit{sequential task} RL. In typical parallel task training \citep{yu_metaworld19}, a new MDP is sampled from $\pr_\mathcal M$ at the start of every episode and is associated with a particular input feature $g\in\mathcal G$ that indicates to the agent which task has been sampled. The agent's performance is evaluated on all tasks $M\in\mathcal M$ together. 
In the \textit{sequential} task setting \citep{Moskovitz:2021_rpotheory,pacchiano2022joint}, tasks (MDPs) are sampled one at a time from  $\pr_\mathcal M$, with the agent training on each until convergence. In contrast to continual learning \citep{parkerholder2021unclear}, the agent's goal is simply to learn a new policy for each task more quickly as more are sampled, rather than learning a single policy which maintains its performance across tasks. 
Another important setting is \textit{meta-RL}, which we do not consider here. In the meta-RL setting, the agent trains on each sampled task for only a few episodes each with the goal of improving few-shot performance and is meta-tested on a set of held-out tasks  \citep{yu_metaworld19,finn2017model}. 

\paragraph{Regularized Policy Optimization}
One common approach which has been shown to to improve performance is \textit{regularized policy optimization}  \citep[RPO;][]{Schulman:2017,Schulman:2018_entropy,Levine:2018_rlai,Agarwal:2020_pg_theory,Pacchiano:2019,tirumala2020behavior,Abdolmaleki:2018}. In \textsc{RPO}, a convex regularization term $\Omega(\theta)$ is added to the objective: $\mathcal J^{\mathrm{RPO}}_\lambda(\theta) = V^{\pi_\theta} - \lambda \Omega(\theta)$. In the single-task setting, the regularization term is often used to approximate trust region \citep{Schulman:2015}, proximal point \citep{Schulman:2017}, or natural gradient \citep{Kakade:2002,Pacchiano:2019,Moskovitz:2021} optimization, or to prevent premature convergence to local maxima \citep{haarnoja_2018sac,lee2018sparse}. 

In multitask settings, the regularization term for \textsc{RPO} typically takes the form of a divergence measure penalizing the policy responsible for taking actions $\pi_\theta$, which we'll refer to as the \textit{control policy}, for deviating from some \textit{default policy} $\pi_w$, which is intended to encode generally useful behavior for some family of tasks \citep{Teh:2017_distral,Galashov:2019,Goyal:2019,Goyal:2020,Moskovitz:2021_rpotheory}. The intuition behind such approaches is that by capturing behavior which is on average useful for some family of tasks, $\pi_w$ can provide a form of beneficial supervision to $\pi_\theta$ when obtaining reward from the environment is challenging, either because $\pi_\theta$ has been insufficiently trained or rewards are sparse. \cite{Moskovitz:2021_rpotheory} took a step towards formalizing this intuition, demonstrating that using a default policy which is in expectation sufficiently ``close'' to the optimal policies for a distribution of tasks can improve convergence rates on new tasks. Popular methods for constructing the default policy include marginalizing over goal-specific policies in multi-goal settings, i.e., $\sum_{g\in\mathcal G} P(g) \pi_\theta(a|s,g)$ \citep{Goyal:2019} or distillation ($\argmin_w\kl[\pi_\theta(a|s), \pi_w(a|s)]$) \citep{Teh:2017_distral,Galashov:2019}. 


\section{The Minimum Description Length Principle}
\label{sec:mdl-principle}

\paragraph{General principle} Simply storing a representation of all environment interactions across multiple tasks is computationally infeasible, and so multitask \textsc{RPO} algorithms offer a compressed representation in the form of a default policy. However, the type of information which is compressed (and that which is lost) is often hard-coded \textit{a priori}. Preferably, we'd like an approach which can distill structural regularities among tasks without needing to know what they are beforehand. The \textit{minimum description length} (MDL) framework offers a principled approach to this problem. So-called ``ideal'' MDL seeks to find the shortest solution written in a general-purpose programming language\footnote{The \textit{invariance theorem} \citep{kolmogorov_invariance65} ensures that, given a sufficiently long sequence, Kolmogorov complexity is invariant to the choice of general-purpose language.} which accurately reproduces the data---an idea rooted the concept of Kolmogorov complexity \citep{li-vitanyi_kolmo97}. Given the known impossibility of computing Kolmogorov complexity for all but the simplest cases, a more practical MDL approach instead prescribes selecting the hypothesis $H^\star$ from some hypothesis class $\mathcal H$ which minimizes the two-part code 
\begin{align} \label{eq:mdl}
  H^\star = \argmin_{H\in\mathcal H} L(\mathcal D|H) + L(H), 
\end{align}
where $L(\mathcal D|H)$ is the number of bits required to encode the data given the hypothesis and $L(H)$ is the number of bits needed to encode the hypothesis itself. There are a variety of so-called \textit{universal} coding schemes which can be used to model \cref{eq:mdl}. 

\paragraph{Variational code} One popular encoding scheme is the variational code~\citep{blier2018description, hinton1993keeping, honkela2004variational}:
\begin{equation}
  L_\nu^\text{var}(\mathcal{D}) = 
  \underbrace{\E_{\theta \sim \nu} \left [ - \log p_\theta(\mathcal{D})\right ]}_{L^{\text{var}}(\mathcal{D} | H)} + \underbrace{\DKL{\nu(\cdot)}{p(\cdot)}}_{L^{\text{var}}(H)}
  \label{eq:variational_code}
\end{equation}
where the hypothesis class is of a set of parametric models $\mathcal{H} = \{ p_{\theta}(\mathcal{D}) : \theta \in \Theta\}$. 
The model parameters are random variables with prior distribution $p(\theta)$ and $\nu(\theta)$ is any distribution over $\Theta$.
Minimizing $L_\nu^\text{var}(\mathcal D)$ with respect to $\nu$ is equivalent to performing variational inference, maximizing a lower-bound to the data log-likelihood $\log p(\mathcal{D})  = \log \int p(\theta) p_\theta(\mathcal{D}) d \theta \geq -L^{\text{var}}_\nu(\mathcal{D})$. Roughly speaking, MDL encourages the choice of ``simple'' models when limited data are available \citep{grunwald_mdl2004}. In the variational coding scheme, simplicity is enforced via the choice of prior. 

\paragraph{Sparsity-inducing priors and variational dropout} %
Choosing sparsity-inducing priors is a fundamental way to improve the compression rate within the variational coding scheme, as such priors encourage the model to prune out parameters that do not contribute to reducing $L^\text{var}(\mathcal{D} | \theta)$.
Many sparsity-inducing priors belong to the family of scale mixtures of normal distributions~\citep{andrews1974scale}:
\begin{align}
  z  \sim p(z), \quad \theta \sim p(\theta | z ) = \mathcal{N}(w; 0, z^2)
\end{align}
where $p(z)$ defines a distribution over the variance $z^2$.
Common choices of $p(z)$ include the Jeffreys prior $p(z) \propto |z|^{-1}$~\citep{jeffreys_prior46}, the inverse-Gamma distribution, and the half-Cauchy distribution~\citep{polson2012half, gelman2006prior}. Such priors have deep connections to MDL theory. For example, the Jeffreys prior in conjunction with an exponential family likelihood is asymptotically identical to the \textit{normalized maximum likelihood} estimator, perhaps the most fundamental `MDL' estimator \citep{grunwald_mdl2019}. 

\textit{Variational dropout} (VDO) is an effective algorithm for minimizing \Cref{eq:variational_code} for these sparsity-inducing priors~\citep{louizos2017bayesian,kingma2015variational,molchanov2017variational}.
Briefly, this involves choosing an approximate posterior distribution with the form
\begin{align}
  p(w, z|\mathcal{D}) \approx \nu(w, z) = \mathcal{N}(z; \mu_z, \alpha \sigma^2_z ) \mathcal{N}(w; z\mu, z^2 \sigma^2 I_d)
\end{align}
and optimizing \Cref{eq:variational_code} via stochastic gradient descent on the variational parameters given by $\{ \alpha, \mu_z, \sigma^2_z, \mu, \sigma^2 \}$. As its name suggests---and importantly for its ease of application to large models---VDO can be implemented as a form of dropout \citep{srivastava:14_dropout} by reparameterizing the noise on the weights as activation noise \citep{kingma2015variational}. 
Application of VDO to Bayesian neural networks has achieved impressive compression rates, sparsifying deep neural networks while maintaining prediction performance on supervised learning problems~\citep{molchanov2017variational,louizos2017bayesian}. Equipped with a powerful approach for MDL-grounded posterior inference, we can now integrate these ideas with multitask \textsc{RPO}.

\section{Minimum Description Length Control}
As part of its underlying philosophy, the MDL principle holds that 1) \textit{learning} is the process of discovering regularity in data, and 2) any regularity in the data can be used to \textit{compress} it \citep{grunwald_mdl2004}. Applying this perspective to RL is non-obvious---from the agent's perspective, what `data' is it trying to compress? Our hypothesis, which forms the basis for the framework we propose in this paper, is that an agent faced with a set of tasks in the world should seek to elucidate structural regularity from the environment interactions generated by the optimal policies for the tasks. 
This makes intuitive sense: the agent ought to compress information which indicates how to correctly perform the tasks with which it is faced. 
That is, we propose that the \textit{data} in multitask RL are the state-action interactions generated by the optimal policies for a set of tasks: 
$
    \mathcal D = \left \{\mathcal D_M \right \}_{M\in\mathcal M} = \left \{ (s, a) \ : \ \forall s\in\St, a\sim\pi^\star_M(\cdot|s)  \right \}_{M\in\mathcal M}
$
This interpretation is in line with work suggesting that a useful operational definition of `task' can be derived directly from the set of optimal (or near-optimal) policies it induces \citep{Abel:2021_reward}. 

Importantly, this interpretation also suggests a natural mapping to the multitask \textsc{RPO} framework. In this view, the control policy is responsible for learning and the default policy for compression: by converging to the optimal policy for a given task, the control policy ``discovers'' regularity which is then distilled into a low-complexity representation by the default policy. 
In our approach, the default policy is encouraged to learn a compressed representation not by artificially constraining the network architecture or via hand-designed information asymmetry, but rather through a prior distribution $p(w)$ over its parameters which biases a variational posterior $\nu(w)$ towards simplicity. 
The default policy is therefore trained to minimize the variational code:
%
\begin{align}
\begin{split}
  \argmin_{\nu\in\mathsf N}
  \E_{\stackrel{s,a\sim \mathcal D}{w\sim\nu}} & -\log \pi_w(a|s) + \kl[\nu(\cdot), p(\cdot)]
  \\ &=
  \argmin_{\nu\in\mathsf N}
  \E_{M\sim\pr_\mathcal M}\E_{\stackrel{s\sim d^{\pi^\star_M}}{w\sim\nu_\phi}}\kl[\pi_M^\star(\cdot|s), \pi_w(\cdot|s)] + \kl[\nu(\cdot), p(\cdot)],
\end{split}
\label{eq:elbo}
\end{align}
%
where $\mathsf N$ is the distribution family for the posterior.
Taken together, this suggests the iterative multitask algorithm presented in \cref{alg:mdlc_replay}, in which for each round $k$, a new task $M_k$ is sampled, the control policy $\pi_\theta$ is trained to approximate the optimal policy $\pi^\star_k$ via RPO, and the result is compressed into a new default policy distribution $\nu_{k+1}$. In the following sections, we further motivate sparsity-inducing priors for the default policy in multitask settings, derive formal performance guarantees for  \textsc{MDL-C}, and demonstrate its empirical effectiveness.

\begin{algorithm}[!t]
	\caption{MDL-C for Sequential Multitask Learning with Persistent Replay}\label{alg:mdlc_replay}
		\begin{algorithmic}[1] 
		    \STATE require: task distribution $\pr_\mathcal M$, policy class $\Theta$, non-increasing coefficients $\{\eta_{k}\}_{k=1}^K$
		    \STATE initialize: default policy distribution  $\nu_1\in\mathsf{N}\subseteq\mathcal P(\Theta)$, default policy dataset $\mathcal D_0 \gets \emptyset$
			\FOR{tasks $k=1, 2, \dots, K $}
			    \STATE Sample a task $M_k=(\St, \A, \mathsf P_k, r_k, \gamma_k, \rho_k)\sim \mathcal P_\mathcal M(\cdot)$
                \STATE Optimize control policy: 
                \begin{equation} 
                    \theta_k \gets \argmax_{\theta\in\Theta} V^\pi_{M_k} - \alpha \E_{s\sim d^\pi_{\rho_k}}\E_{w\sim\nu_k}\kl[\pi_w(\cdot|s), \pi_\theta(\cdot|s)]
                \end{equation}
                \STATE Add data to default policy replay:
                \begin{equation}
                    \mathcal D_k \gets \mathcal D_{k-1} \cup \{(s_m, \hat\pi_{\theta_k}(s_m))\}_{m=1}^M,
                \end{equation}
                where $M = |\St|$ for finite/small state spaces
                \STATE 
                    Update default policy distribution:
                    \begin{align}
                        \nu_{k+1} \gets \argmin_{\nu \in \mathsf N} \frac{1}{\eta_{k-1}} \kl[\nu(\cdot), p(\cdot)] + \sum_{i=1}^k \sum_{m=1}^M\E_{w\sim\nu}\kl[\hat\pi_{\theta_i}^\star(\cdot|s_m), \pi_w(\cdot|s_m)]
                    \end{align}
                
			 \ENDFOR
	\end{algorithmic}
\end{algorithm}

\subsection{Motivating the choice of sparsity-inducing priors}
\label{subsec:sparsity}
In \Cref{sec:mdl-principle}, compression (via pruning extraneous parameters) is the primary motivation for using sparsity-inducing priors $p(w)$ that belong to the family of scaled-mixtures of normal distributions. Intuitively, placing a distribution over the default parameters reflects the agent's \textit{epistemic uncertainty} about the task distribution---when few tasks have been sampled, a sparse prior prevents the default policy from overfitting to what may ultimately be spurious correlations in the limited data that the agent has collected.
Here, we make this motivation more precise, describing an example generative model of optimal policy parameters which provides a principled interpretation for prior choice $p(z)$ in multitask RL.

\begin{figure}[t!]
\centering
\includegraphics[width=0.65\linewidth]{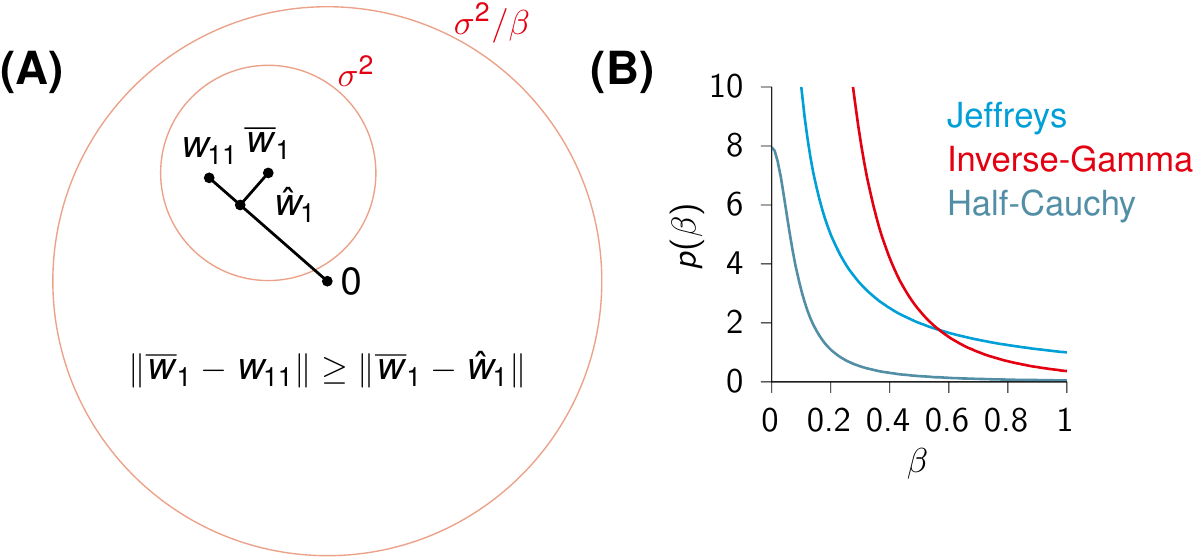} 
\caption{
    \label{fig:schematic}
    \textbf{(A)} Illustration of a generative model of optimal policy parameters. $\hat{w}_1=(1-\beta) w_{11}$ shrinks towards the origin, becoming a closer estimate of $\overline{w}_1$ than $w_{11}$.
    \textbf{(B)} Sparsity-inducing priors over $\beta$.
}
\end{figure}

\paragraph{Generative model of optimal policy parameters} Consider a set of tasks $\mathcal{M} = \{ M_{ik} \}_{i=1,k=1}^{I, K_i}$ that are clustered into $I$ groups, 
such that the MDPs in each group are more similar to one another than to members of other groups. As an example, the overall family $\mathcal M$ could be all sports, while clusters $\mathcal M_i\subseteq \mathcal M$ could consist of, say, ball sports or endurance competitions. 
To make this precise, we assume that the optimal policies of every MDP belong to a parametric family $\Pi =\{ \pi_w(\cdot|s) : w \in \mathbb{R}^d, \forall s \in \mathcal{S} \}$ (e.g., softmax policies with parameters $w$), and that the optimal policies for each group are randomly distributed within parameter space.
In particular, we assume that the parameters of the optimal policies of $\mathcal{M}$ have the following generative model:
\begin{align*}
    \overline{w}_i | \beta, \sigma^2 \sim \mathcal{N}\left(\overline{w}_m; 0, (1-\beta)\beta^{-1}\sigma^2 I_d\right), \quad 
    w_{ik} | \overline{w}_i, \sigma^2 \sim \mathcal{N}\left(w_{ik}; \overline{w}_i, \sigma^2 I_d\right).
\end{align*}
where $I_d$ is the $d-$dimensional identity matrix.
If we marginalize out $\overline{w}_i$, we get the marginal distribution $p(w_{ik} | \beta, \sigma^2) = \mathcal{N}(w_{ik}; 0, \sigma^2\beta^{-1} I_d)$. We can therefore visualize the parameter distribution of the optimal policies for $\mathcal M$ as a $d$-dimensional Gaussian within which lie clusters of optimal policies for related tasks which are themselves normally distributed (see \cref{fig:schematic}A for a visualization of $d=2$). 

\paragraph{Interpretation of {$\beta$}{}} The parameter $\beta \in (0,1]$ has the following interpretation (see \Cref{fig:schematic}A):
\begin{equation*}
  \beta = \frac{\text{squared distance between optimal policy parameters within a group}}{
    \text{squared distance between optimal policies in }\mathcal{M}}.
\end{equation*}
Intuitively, $\beta$ determines how much information one gains about the optimal parameters of a task in a group, given knowledge about the optimal parameters of another task in the same group.
To see this, we compute our posterior belief about the value $\overline{w_i}$ given observation of $w_{ik}$:
\begin{align*}
  p(\overline{w}_i | w_{ik}, \beta, \sigma^2) = \mathcal{N}\left (\overline{w}_i; (1-\beta)w_{ik}, (1-\beta)\sigma^2 I_d \right ).
\end{align*}

\noindent When $\beta=1$ (inner circle in \Cref{fig:schematic}A has the same radius as the outer circle), our posterior mean estimate of $\overline{w}_i$ is simply $0$, suggesting we have learned nothing new about the mean of the optimal parameters in group $i$, by observing $w_{ik}$.
In the other extreme when $\beta \rightarrow 0$, the posterior mean approaches the maximum-likelihood estimator $w_{ik}$, suggesting that observation of $w_{ik}$ provides maximal information about the optimal parameters in group $i$.
Any $\beta$ in between the two extremes results in an estimator that ``shrinks'' $w_{ik}$ towards $0$. The value of $\beta$ thus has important implications for multitask learning. 
Suppose an RL agent learns the optimal parameters $w_{11}$ (task 1, group 1), and proceeds to learn task $2$ in group $1$.
The value of $\beta$ determines whether $w_{11}$ can be used to inform the agent's learning of $w_{21}$. In this way, $\beta$ determines the effective degree of epistemic uncertainty the agent has about the task distribution. 

\paragraph{Choice of \texorpdfstring{$p(\beta)$}{} and connection to \texorpdfstring{$p(z)$}{}} %
The importance of $\beta$ thus raises the question: what should $\beta$ be?
As any good Bayesian would do, instead of treating $\beta$ as a parameter, we can choose a prior $p(\beta)$ and perform Bayesian inference. Ideally, $p(\beta)$ should (i) encode our prior belief about the extent to which the optimal parameters cluster into groups and (ii) result in a posterior mean estimator $\hat{w}^{(p(\beta))}(x) = 1 - \E \left [ \beta | x \right ] x$ that is close to $\overline{w}$ for $x | \overline{w} \sim \mathcal{N}(x; \overline{w}, \sigma^2)$. 
This condition encourages the expected default policy (under the posterior $\nu$; \Cref{eq:elbo}) to be close to optimal policies in the same MDP group (centered at $\overline{w}$). One prior choice that satisfies both conditions is  $p(\beta) \propto \beta^{-1}$.
It places high probability for small $\beta$ and low probability for high $\beta$, thus encoding the prior belief that the optimal task parameters are clustered~(see \Cref{fig:schematic}B; blue).
It is instructive to compare $p(\beta) \propto \beta^{-1}$ with two extreme choices of $p(\beta)$.
When $p(\beta) = \delta(\beta-1)$, $p(z) = \delta(\sigma)$ and the marginal $p(w)$ is the often-used Gaussian prior over the parameters $w$ with fixed variance $\sigma^2$. 
This corresponds to the prior belief that knowing $w_{i1}$ provides no information about $w_{i2}$.
On the other hand,  $p(\beta)=\delta(\beta)$ recovers a uniform  prior over the parameters $w$ and reflects the prior belief that the MDP groups are infinitely far apart.
In relation to (ii), one can show the $\hat{w}^{(p(\beta))}$ strictly dominates the maximum-likelihood estimator $\hat{w}^{(\text{ML})}(x) = x$~(\citealp{efron1973stein}; \Cref{supp:sec:stein}),
for $p(\beta) \propto \beta^{-1}$. 
This means $\text{MSE}(\overline{w}, \hat{w}^{(p(\beta))}) \leq \text{MSE}(\overline{w}, \hat{w}^{(\text{ML})})$ for all $\overline{w}$, where $\text{MSE}(\overline{w}, \hat{w}) = \E_{x \sim \mathcal{N}(x; \overline{w}, \sigma^2)} \|\overline{w} - \hat{w}(x)\|^2$.

\paragraph{Connection to \texorpdfstring{$p(z)$}{} and application of VDO} Defining $z^2=\sigma^2 \beta^{-1}$ and applying the change-of-variable formula to $p(\beta) \propto \beta^{-1}$ gives $p(z) \propto |z|^{-1}$ and thus the Normal-Jeffreys prior in \Cref{sec:mdl-principle}.
This correspondence enables the application of VDO (see \Cref{sec:mdl-principle}) to obtain an approximate posterior $\nu(w, z)$ which minimizes the variational code~\Cref{eq:elbo}.
Similar correspondences may also be derived for the inverse-Gamma distribution and the half-Cauchy distribution, which both satisfy (i) and (ii)~(see \Cref{fig:schematic}B; \cref{supp:sec:stein}).

\subsection{Performance Analysis}
At a fundamental level, we'd like assurance (i) that  \textsc{MDL-C}'s default policy will be able to effectively distill the optimal policies for previously observed tasks, and (ii) that regularization using this default policy gives strong performance guarantees for the control policy on future tasks. 

\paragraph{Performance Characteristics} One way we can verify (i) is to obtain an upper bound on the average KL between default policies sampled from the default policy distribution and an optimal policy for a task sampled from the task distribution. An important feature of  \textsc{MDL-C} is that each term in the objective function which depends directly on the default policy distribution is convex with respect to it. This enables us to analyze the properties of the learned default policy distribution through the lens of \textit{online convex optimization} (OCO). In OCO, the learner observes a series of convex loss functions $\ell_k: \mathsf N \to \reals$, $k=1,\dots,K$, where $\mathsf N \subseteq \reals^d$ is a convex set. After each round, the learner produces an output $x_k \in \mathsf N$ for which it will then incur a loss $\ell_k(x_k)$ \citep{orabona_online19}. At round $k$, the learner is usually assumed to have knowledge of $\ell_1, \dots, \ell_{k-1}$, but no other assumptions are made about the sequence of loss functions. The learner's goal is to minimize its average regret. For further background on OCO, see \cref{sect:oco}.  
One OCO algorithm which enjoys sublinear regret is \textit{follow the regularized leader} (FTRL). In each round of FTRL, the learner selects the solution $x\in\mathsf N$ according to the following objective:
$
  x_{k+1} = \argmin_{x\in\mathsf N} \psi_k(x) + \sum_{i=1}^{k-1}\ell_i(x),
$
where $\psi: \mathsf N\to\reals$ is a convex regularization function. 
We can now show that  \textsc{MDL-C} objective for the default policy distribution can be viewed as an implementation of FTRL. To see this, note that by setting $x_k = \nu_k$, $\psi_k(\nu) = \kl[\nu, p]$, and $\ell_k(\nu) = \E_{w\sim\nu} \kl[\pi_w, \pi_k^\star]$, we recover the procedure in \cref{alg:mdlc}. Using standard results from OCO, this connection allows us to bound  \textsc{MDL-C}'s regret in learning the default policy distribution. All proofs are provided in \cref{sect:performance_proofs}.
%

\begin{restatable}[Persistent Replay FTRL Regret; \citep{orabona_online19}, Corollary 7.9]{proposition}{persistentregret} \label{thm:persistent_regret}
  Let tasks $M_k$ be independently drawn from $\pr_\mathcal M$ at every round, and let them each be associated with a deterministic optimal policy $\pi_k^\star: \St \to \A$.  
  We make the following mild assumptions: i) $\pi_w(a^\star|s) \geq \epsilon > 0$ $\forall s\in\St$, where $a^\star=\pi_k^\star(s)$ and $\epsilon$ is a constant. ii) $\min_\nu \kl[\nu(\cdot), p(\cdot)] = 0$ asymptotically as $\mathrm{Var}[\nu] \to \infty$. 
  Then with $\eta_{k-1} = \log(1/\epsilon)\sqrt{k}$, \cref{alg:mdlc_replay} guarantees 
  \begin{align} \label{eq:persistent_pi0_ub}
    \frac{1}{K} \sum_{k=1}^K \ell_k(\nu_k)  - \frac{1}{K}\sum_{k=1}^K \ell_k(\bar\nu_K) \leq \left(\kl[\bar\nu_K, p] + 1 \right)\frac{\log(1/\epsilon)}{\sqrt{K}},
\end{align}
  where $\bar\nu_K = \argmin_{\nu\in\mathsf N} \sum_{k=1}^K \ell_k(\nu)$. 
\end{restatable}

Intuitively, this result shows that the average regret is upper-bounded by factors which depend on the divergence of the barycenter distribution from the prior and the ``worst-case'' prediction of the default policy. Crucially, we can see that the average regret is $\mathcal O(1/\sqrt{K})$: the KL between the default policy distribution and the barycenter distribution goes to zero as the number of tasks $K\to\infty$. 

Importantly, we can also now be assured of point (ii) above, in that this result can be used to obtain a sample-complexity bound for the \textit{control} policy. Specifically, we can use \cref{thm:default_regret} to place an upper-bound on the total variation distance between default policies sampled from $\nu$ and the KL between the maximum likelihood solution and a sparsity-inducing prior $p$. This is useful, as it allows to translate low regret for the default policy into a sample complexity result for the control policy using \cite{Moskovitz:2021_rpotheory}, Lemma 5.2.
\begin{restatable}[Control Policy Sample Complexity]{proposition}{controlperf} \label{thm:controlperf}
  Under the setting described in \cref{thm:default_regret}, denote by $T_k$ the number of iterations to reach $\epsilon$-error for $M_k$ in the sense that $
        \min_{t\leq T_k} \{V^{\pi_k^\star} - V^{(t)}\} \leq \eps.
    $
    Further, denote the upper-bound in \cref{eq:ub_pi0} by $G(K)$. In a finite MDP, from any initial $\theta^{(0)}$, and following gradient ascent, $\expect[M_k\sim\mathcal P_\mathcal M]{T_k}$ satisfies:
    \begin{align*}
        \expect[M_k\sim\mathcal P_{\mathcal M_i}]{T_k} \geq  \frac{80\vert\mathcal{A} \vert^2 \vert \mathcal{S}\vert^2}{\epsilon^2(1-\gamma)^6} \mathbb E_{\substack{M_k\sim\mathcal P_{\mathcal M_i} \\
        s\sim\Unif_\St}}\left[\kappa_\A^{\alpha_k}(s) \left\| \frac{d_{\rho}^{\pi_k^*} }{\mu}  \right \|_\infty^2\right],
    \end{align*}
    where $\alpha_k(s) \coloneqq \tv(\pi_k^\star(\cdot|s), \hat{\pi}_0(\cdot|s)) \leq \sqrt{G(K)}$, $\kappa_\A^{\alpha_k}(s) = \frac{2|\A|(1 - \alpha(s)) }{2|\A|(1 - \alpha(s)) - 1}$, and $\mu$ is a measure over $\St$ such that $\mu(s)>0$ $\forall s\in\St$. 
\end{restatable}
The core takeaway from these results is that as the agent is trained on more tasks, the default policy distribution regret, upper-bounded by $G(K), $decreases asymptotically to zero, and as the default policy regret decreases, the control policy will learn more rapidly, as $\mathrm{poly}(G(K))$.

\section{Experiments} \label{sect:experiments}
We tested the  \textsc{MDL-C} framework empirically in two different settings: 1) multitask learning with on-policy control and a discrete action space and 2) meta-learning with off-policy, continuous control. Our objective is to empirically test the multitask learning benefits of  \textsc{MDL-C}. To quantify performance, in addition to measuring per-task reward, we also report the cumulative regret for each method in each experimental setting in  \cref{table:4rooms}.

\subsection{2D Navigation}
\begin{figure*}[!t]
  \centering
  \includegraphics[width=0.99\textwidth]{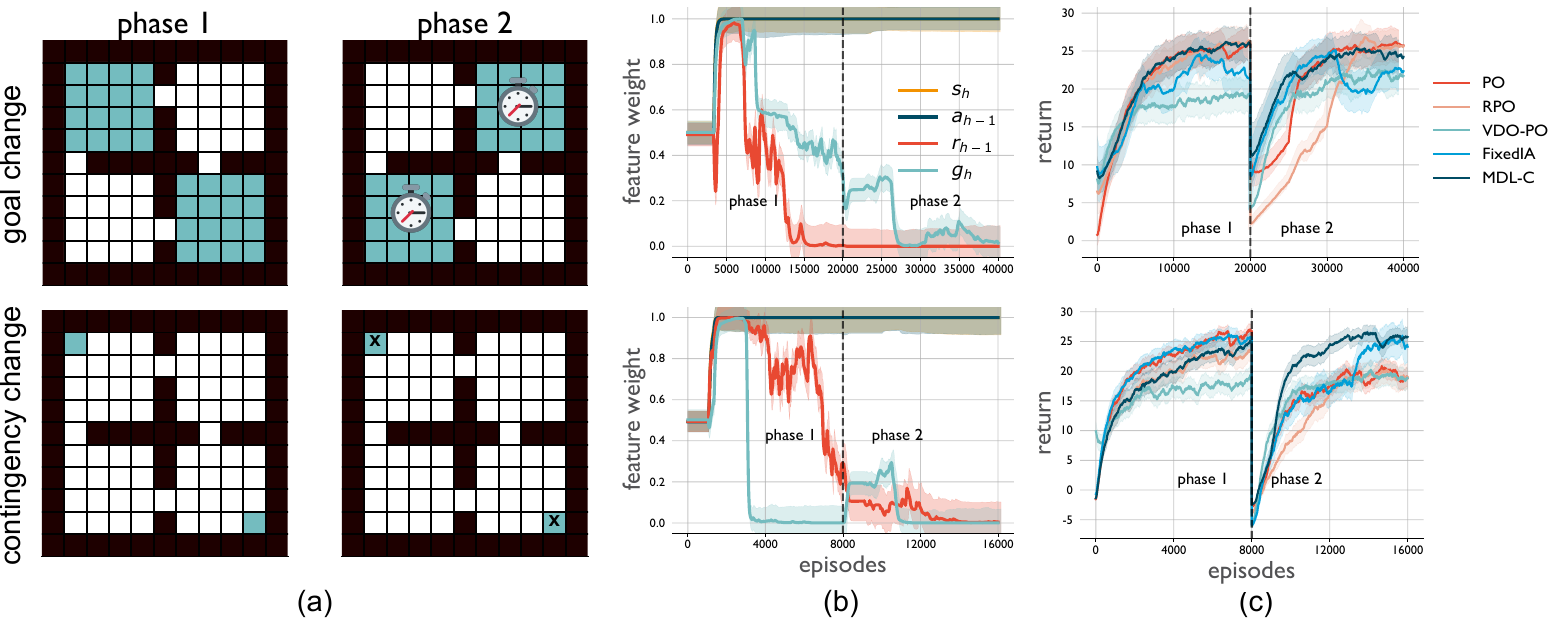}
  \caption{MDL-C rapidly adapts to new goal locations (top row) and rule changes (bottom row). All curves represent averages taken over 10 random seeds, with the shading indicating standard error.}
  \label{fig:4rooms}
\end{figure*}

We first test  \textsc{MDL-C} on 2D navigation in the classic \textsc{FourRooms} environment (\cref{fig:4rooms}a, \citep{Sutton:1991_fourrooms}). The baselines in this case are \textsc{PO} (entropy-regularized policy optimization), \textsc{RPO} (regularized policy optimization with no constraint on the default policy), \textsc{VDO-PO} (an agent whose control policy is directly regularized without a default policy), and \textsc{ManualIA} (the agent from \cite{Galashov:2019} in which the goal feature is manually witheld from the default policy). As input, the agent receives a 16-dimensional vector containing the index of the current state, a flattened $3\times 3$ local view of its surrounding environment, its previous action taken encoded as a 4-dimensional one-hot vector, the reward on the previous timestep, and a feature indicating the goal state index. The base learning algorithm in all cases is advantage actor critic (A2C; \citep{Mnih:2016_a3c}). Further experimental details can be found in \cref{sect:expt_details}.

\paragraph{Generalization Across Goals}
In the first setting, we test  \textsc{MDL-C}'s ability to facilitate rapid learning on previously unseen goals. In the first phase of training, a single goal location is randomly sampled at the start of each episode, and may be placed anywhere in two of the four rooms in the environment (\cref{fig:4rooms}a, top left). In the second phase of training, the goal location is again randomly sampled at the start of each episode, but in this case, only in the rooms which were held out in the first phase. Additionally, the agent is limited to 25 rather than 100 steps per episode. Each phase comprises 20,000 episodes, and in each phase, the agent may start each episode anywhere in the environment. Importantly, VDO induces the  \textsc{MDL-C} default policy to ignore input features which are, on average, less predictive of the control policy's behavior. In this case, the default policy learns to ignore the goal feature and the reward obtained on the previous timestep. This is because, when averaging across goal locations, the agent's current position ($s_h$) and the direction in which it was last heading ($a_{h-1}$) are more informative of its next action---typically, heading towards the nearest door. In contrast, the un-regularized default policy of the \textsc{RPO} agent does not drop these features (\cref{sect:add_figs} for a visualization and \cref{sect:expt_details} for more details). By learning to ignore the specific goals present in phase 1 and encoding behavior that is useful independent of goal location,  \textsc{MDL-C}'s default policy makes a more effective regularizer in the phase 2, enabling the control policy to adapt more quickly than other methods (\cref{fig:4rooms}c, top), particularly \textsc{RPO}, which overfits to phase 1's goals. \textsc{ManualIA} also adapts quickly, as its default policy is hard-coded to ignore the goal feature.  

\paragraph{Robustness to Rule Changes}
In this setting, we again split training into two phases, in this case each consisting of 8,000 episodes. There are only two possible goal locations, one at the top left of the environment, and the other at the bottom right, with one goal randomly sampled at the start of each episode. In phase 1 of training, the agent receives a goal feature as input which indicates the state index of the rewarded location for that episode. In phase 2, however, the goal feature switches from marking the reward location to marking the unrewarded location. That is, if the reward is in the top left, the goal feature will point to the bottom right. In this setting, the danger for the agent isn't overfitting to a particular goal or goals, but rather ``overfitting'' to the reward-based rules associated with a given feature. As we saw in \cref{fig:4rooms}c (top), an un-regularized default policy, will simply copy the control policy and overfit to a particular setting. Once again, however, the  \textsc{MDL-C} default policy learns to ignore features which are, on average, less useful for predicting the control policy's behavior---the goal and previous reward features. This renders the agent more robust to contingency switches like the one described, as we can see in \cref{fig:4rooms}c (bottom). 
These examples illustrate that MDL-C enables agents to effectively learn the consistent structure of a group of tasks, regardless of its semantics, and ``compress out'' information which is less informative on average.

\begin{figure*}[!t]
  \centering
  \includegraphics[width=0.99\textwidth]{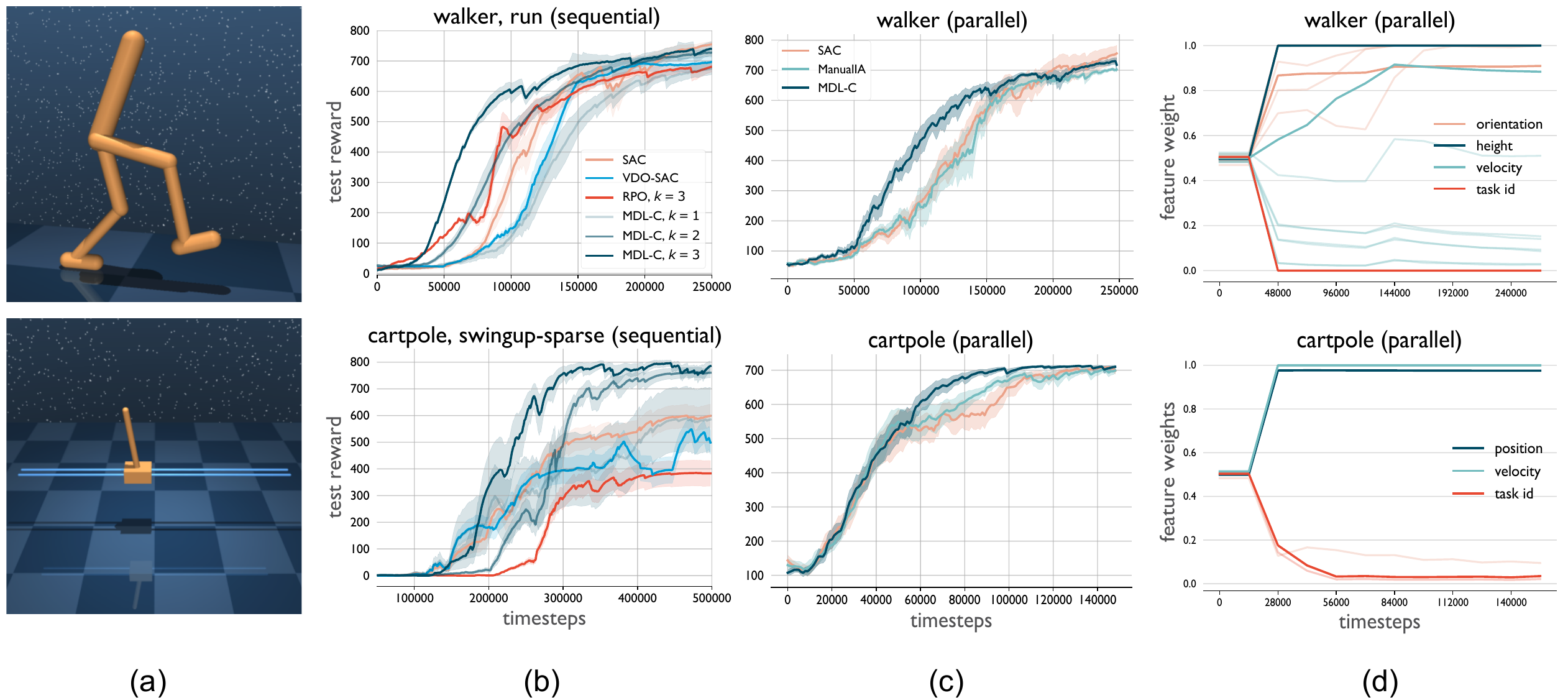}
  \caption{MDL-C improves both sequential and parallel learning in continuous control tasks. All curves represent averages taken over 8 random seeds, with the shading indicating standard error. In (d), solid curves represent averages over each feature within a category.}
  \label{fig:dmc}
\end{figure*}

\subsection{Continuous Control}
A more challenging application area is that of high-dimensional continuous control. To test  \textsc{MDL-C}'s performance in this setting, we presented agents with multitask learning problems using environments from the DeepMind Control Suite (DMC; \citep{tassa_2018DMC}). We used soft actor critic (\text{SAC}; \citep{haarnoja_2018sac}) as the base agent. We tested \textsc{MDL-C} on two separate multitask paradigms: sequential tasks and parallel tasks on two domains from DMC: walker and cartpole (\cref{fig:dmc}a). Additional training details can be found in \cref{sect:expt_details}. 

\paragraph{Sequential Tasks} In the sequential task setting, tasks are sampled one at a time uniformly without replacement from the available tasks within each domain, with the default policy distribution $\nu$ conserved across tasks. The agent's objective is to accelerate learning on each successive task, as measured by cumulative regret.  For walker, these tasks are \texttt{stand},  \texttt{walk}, and \texttt{run}. In \texttt{stand}, the agent is rewarded for increasing the height of its center of mass, and in the latter two tasks, an additional reward is given for forward velocity. For cartpole, there are four tasks: \texttt{balance}, \texttt{balance-sparse}, \texttt{swingup}, and \texttt{swingup-sparse}. In the \texttt{balance} tasks, the agent must keep a rotating pole upright, and in the \texttt{swingup} tasks, it must additionally learn to swing the pole upwards from an initial downward orientation. Performance results for the hardest task within each domain (\texttt{run} in walker and \texttt{swingup-sparse} in cartpole) for each method are plotted in \cref{fig:dmc}b, where $k$ indicates the task round at which the task was sampled. We can see that as $k$ increases in both cases (as more tasks have been seen previously), \textsc{MDL-C}'s performance improves substantially. Importantly, the \textsc{RPO} agent's default policy, which is un-regularized, overfits to the previous task, essentially copying the optimal policy's behavior. This can severely hinder the agent's performance when the subsequent task requires different behavior. For example, on \texttt{swingup-sparse}, if the previous task is \texttt{swingup}, the \textsc{RPO} agent performs very well, as the goal is identical. However, if the previous task is \texttt{balance} or \texttt{balance-sparse}, the agent never learns to swing the pole upwards, significantly reducing the resulting average performance. 

\paragraph{Parallel Tasks} We also tested parallel-task versions of \textsc{SAC}, \textsc{ManualIA}, and \textsc{MDL-C} based on the model of \cite{yu_metaworld19}. In this framework, a task within each domain is randomly sampled at the start of each episode---the task for each episode is communicated to the agent via a one-hot ID feature---and the agent aims to learn a single control policy that can perform well on all tasks within the domain. The performance of each agent is plotted in \cref{fig:dmc}c, where we can again see that \textsc{MDL-C} accelerates convergence relative to the baseline methods. This marks a difference compared to the easier FourRooms environment, in which \textsc{MDL-C} and the agent with manual information asymmetry performed roughly the same. As before, one clue to the difference can be found in the input features that the \textsc{MDL-C} default policy chooses to ignore (\cref{fig:dmc}d). For walker, inputs are 24-dimensional, with 14 features related to the joint orientations, 1 feature indicating the height of the agent's center of mass, and 9 features indicating velocity components. For cartpole, there are 5 input dimensions, with 3 pertaining to position and 2 to velocity. In the walker domain, where the performance difference is greatest, the \textsc{MDL-C} agent not only ignores the added task ID feature, but also the several features related to velocity. In contrast, in the cartpole domain, \textsc{MDL-C} only ignores the task ID feature, just as \textsc{ManualIA} does, and the performance gap is smaller. This illustrates that \textsc{MDL-C} learns to compress out spurious information even in settings for which it is difficult to identify \textit{a priori}.

In order to test the effect of the learned asymmetry on performance more directly, we implemented a variant of \textsc{ManualIA} in which all of the features which \textsc{MDL-C} learned to ignore were manually hidden from the default policy (\cref{fig:manual-ia_plus}). Interestingly, while this method improved over standard \textsc{ManualIA}, it didn't completely close the gap with \textsc{MDL-C}, indicating there are downstream effects within the network beyond input processing which are important for the default policy's effectiveness. We hope to explore these effects in more detail in future work.

\section{Related Work} \label{sect:related_work}
    MDL-C can be viewed as an extension of recent approaches to learning default policies (``behavioral priors'') from the optimal policies of related tasks~\citep{Teh:2017_distral, tirumala2020behavior}.
    For a default policy to be useful for transfer learning, it is crucial to balance the ability of the default policy to ``copy'' the control policies with its expressiveness.
    If the default policy is too expressive, it is likely to overfit on past tasks and fail to generalize to unseen tasks.
    Whereas prior work primarily hand-crafts structural constraints into the default policies to avoid overfitting (e.g., by hiding certain state information from the default policy; \citealp{Galashov:2019}), MDL-C learns such a balance from data with sparsity-inducing priors via variational inference. MDL-C may also be derived from the RL-as-inference framework~(\citealp{Levine:2018_rlai}; \Cref{supp:sec:rl_as_inference}).
    MDL-C thus has close connections with algorithms such as MPO~\citep{Abdolmaleki:2018} and VIREL~\citep{Fellows:2020_virel}, discussed in \cref{supp:sec:rl_as_inference}. As a general framework, MDL-C is also connected to the long and well-established literature on choosing appropriate Bayesian priors~\citep{jeffreys_prior46, bernardo2005reference, casella1985introduction}, and more recent work that focuses on learning such priors for large-scale machine learning models~\citep{nalisnick2017learning, nalisnick2021predictive, atanov2018deep}. For a further discussion of related work, particularly concerning the application of MDL to the RL setting, see \cref{sect:more_related}.



\section{Conclusion} \label{sect:conclusion}
Inspired by dual process theories and the MDL principle, we propose a regularized policy optimization framework for multitask RL which aims to learn a simple default policy encoding a low-complexity distillation of the optimal behavior for some family of tasks. By encouraging the default policy to maintain a low effective description length, \textsc{MDL-C} ensures that its default policy does not overfit to spurious correlations among the (approximately) optimal policies learned by the agent. We described \textsc{MDL-C}'s formal properties and demonstrated its empirical effectiveness in discrete and continuous control tasks. 
There are of course limitations of \textsc{MDL-C}, which we believe represent opportunities for future work (see \cref{sect:limits}). In particular, promising research directions include integrating \textsc{MDL-C} with multitask RL approaches which balance a larger set of policies  \citep{Barreto:2020_fastgpi,moskovitz2022_fr,thakoor2022ghms} as well considering nonstationary environments \citep{parkerholder2022accel}.  
We hope \textsc{MDL-C} inspires further on understanding and extending current approaches to multitask RL.

\clearpage
\bibliographystyle{unsrtnat} 
\bibliography{main}

\clearpage

\begin{appendices}

\begin{center}
    {\Large \bfseries Minimum Description Length Control}\\[0.5em]
    {\Large Supplementary Information}
\end{center}
\vspace{1em}

\section{Reinforcement Learning as Inference}
\label{supp:sec:rl_as_inference}
The control as inference framework~\citepapp{Levine:2018_rlai} associates every time step $h$ with a binary ``optimality'' random variable $\mathcal{O}_{h} \in \{0, 1\}$ that indicates whether $a_h$ is optimal at state $s_h$ ($\mathcal{O}_h=1$ for optimal, and $\mathcal{O}_h=0$ for not).
The optimality variable has the conditional distribution $P(\mathcal{O}_h = 1 | s_h, a_h) = \exp(r(s_h, a_h))$, which scales exponentially with the reward received taking action $a_h$ in state $s_h$.

Denote $\bm{\mathcal{O}}_H$ as the event that $\mathcal{O}_{s}=1$ for $s=0, \ldots, H-1$.
Then the log-likelihood that a policy $\pi_w(a|s)$ is optimal over a horizon $H$ is given by:
\begin{align*}
  \pr(\bm{\mathcal{O}}_H)  = 
  \int  \pr(\bm{\mathcal{O}}_H | \tau) 
  \pr^{\pi_w}(\tau | w)  p(w) d\tau dw.
\end{align*}
By performing variational inference, we can lower-bound the log-likelihood with the ELBO:
\begin{align}
  \begin{split}
\log \pr(\bm{\mathcal{O}}_H) \geq
\E_{\nu_{\pi}(\tau)} 
\sum_{h=0}^{H-1}
\left (  r(s_h,a_h) 
- \E_{\nu_\theta(w)}  
\DKL{\pi_\theta(a_h|s_h)}{\pi_w(a_h|s_h)}
\right ) \\
- \DKL{\nu_\phi(w)}{p(w)},
  \end{split}
\label{eq:control-as-inference-objective}
\end{align}
where $\nu_{\theta,\phi}(\tau, w) = \nu_\theta(\tau) \nu_\phi(w)$ is the variational posterior,
\begin{align*}
  \nu_\theta(\tau) &= \rho(s_0) \prod_{h=0}^{H-1} \pr(s_{h+1} | s_h, a_h) \pi_\theta(a_h, s_h)  
\end{align*}
and $\{\theta, \phi\}$ are the variational parameters.
We can maximize this objective iteratively by performing coordinate ascent on $\{\theta, \phi\}$:
\begin{align}
  \theta &\leftarrow \theta + \eta \nabla_\theta 
  \left ( 
  \E_{\nu_{\theta}(\tau)} 
\sum_{h=0}^{H-1}
\left (  r(s_h,a_h) 
- \E_{\nu_\theta(w)}  
\DKL{\pi_\theta(a_h|s_h)}{\pi_w(a_h|s_h)}
\right ) \right ), 
\label{supp:eq:rl_inf_control} \\
  \phi&\leftarrow \phi
  - \eta \nabla_\phi
  \left ( 
  \E_{\nu_{\theta}(\tau)} 
\sum_{h=0}^{H-1}
\E_{\nu_\theta(w)}  
\DKL{\pi_\theta(a_h|s_h)}{\pi_w(a_h|s_h)}
+ \DKL{\nu_\phi(w)}{p(w)}
\right ) 
\label{supp:eq:rl_inf_default}
\end{align}
where $\eta$ is a learning rate parameter.
Note that \Cref{supp:eq:rl_inf_default} is equivalent to \Cref{eq:elbo} and \Cref{algo:eq:default}, and \Cref{supp:eq:rl_inf_control} is equivalent to \Cref{algo:eq:control} with the KL reversed.

\paragraph{Connection to Maximum a Posteriori Policy Optimization (MPO)} %
\textsc{MDL-C} is closely related to MPO~\citep{Abdolmaleki:2018}, with three key differences. 
First, \textsc{MDL-C} performs variational inference on the parameters of the default policy with an approximate posterior $\nu_\phi(w)$, whereas MPO performs MAP inference.
Second, MPO places a normal prior on $w$, which in effect penalizes the L2 norm of $w$.
In contrast, \textsc{MDL-C} uses sparsity-inducing priors such as the normal-Jeffreys prior.
Third, \textsc{MDL-C} uses a parametric $\pi_\theta$, whereas MPO uses a non-parametric one\footnote{In practice, MPO parametrizes $\pi_\theta$ implicitly with a parameterized action-value function and the default policy.}. 
While there is also a parametric variant of MPO, this variant does not maintain $\theta$ and $\phi$ separately. 
Instead, this variant directly sets $\theta$ to $\phi$ in \Cref{supp:eq:rl_inf_control}.
This illustrates the key conceptual difference between \textsc{MDL-C} and MPO. 
\textsc{MDL-C} makes a clear distinction between the control policy $\pi_\theta$ and the default policy $\pi_w$, with the two policies serving two distinct purposes: the control policy for performing on the current task, the default policy for distilling optimal policies across tasks and generalizing to new ones. 
MPO, on the other hand, treats $\pi_\theta$ and $\pi_w$ as fundamentally the same object.

Like MPO, VIREL~\citepapp{Fellows:2020_virel} can be derived from the control as inference framework. 
In fact, \citeauthor{Fellows:2020_virel} showed that a parametric variant of MPO can be derived from VIREL~\citepapp{Fellows:2020_virel}.
The key novelty that sets VIREL apart from both MPO and \textsc{MDL-C} is an adaptive temperature parameter that dynamically updates the influence of the KL term in \Cref{supp:eq:rl_inf_control}.

\section{Additional Related Work} \label{sect:more_related}
Previous work has also applied the MDL principle in an RL context, though primarily in the context of unsupervised skill learning \citep{zhang2021minimum,thrun:94_mdl-skills}. For example, \cite{thrun:94_mdl-skills} are concerned with a set of ``skills'' which are policies defined only over a subset of the state space that are reused across tasks. They consider tabular methods, measuring a pseudo-description length as 
\begin{align}
  DL = \sum_{s\in\St}\sum_{M\in\mathcal M} P^*_M(s) + \sum_{n\in N}|S_n|,
\end{align}
where $P^*_M(s)$ is the probability that no skill selects an action in state $s$ for task $M$ and the agent must compute the optimal $Q$-values in state $s$ for $M$, $N$ is the number of skills, and $|S_n|$ is the number states for which skill $n$ is defined. They then trade off this description length term with performance across a series of tabular environments.

One other related method is \textsc{DISTRAL} \citep{Teh:2017_distral}, which uses the following objective in the parallel task setting:
\begin{align}
  \mathcal J^{\mathrm{DISTRAL}}(\theta, \phi) = V^{\pi_\theta} -  \E_{s\sim d^{\pi_\theta}}\left[\alpha \kl[\pi_\theta(\cdot|s), \pi_\phi(\cdot|s)] + \beta \mathsf H[\pi_\theta(\cdot|s)] \right].
\end{align}
That is, like the un-regularized \textsc{RPO} method, \textsc{DISTRAL} can be seen as performing maximum-likelihood estimation to learn the (unconstrained) default policy, while adding an entropy bonus to the control policy.

\section{Motivating the choice of sparsity-inducing priors}
\label{supp:sec:stein}
As a reminder, the generative model of optimal parameters in \Cref{subsec:sparsity} is given by:
\begin{align}
    \overline{w}_{i} | \beta, \sigma^2 &\sim \mathcal{N}(0, \frac{1-\beta}{\beta} \sigma^2 I_d),\\
    {w}_{ik} | \overline{w}_i, \sigma^2, \beta &\sim \mathcal{N}(\overline{w}, \sigma^2 I_d)
\end{align}
with marginal and posterior densities
\begin{align}
    p(w_{ik} | \sigma^2, \beta) &= \mathcal{N}(0, \sigma^2 \beta^{-1} I_d),\\
    p(\overline{w}_{i} | w_{ik}, \sigma^2, \beta) &= \mathcal{N}\left ((1-\beta) w_{ik}, (1-\beta) \sigma^2 I_d \right ).
\end{align}

In the rest of this section, we set $\sigma^2=1$ for simplicity and drop the indices on $w$ and $\overline{w}$ to remove clutter.

\subsection{Correspondence between \texorpdfstring{$p(z)$}{} and \texorpdfstring{$p(\beta)$}{}}

In \Cref{subsec:sparsity}, we draw a connection between $p(\beta) \propto \beta^{-1}$ and the normal-Jeffreys prior, which is commonly used for compressing deep neural networks~\citepapp{louizos2017bayesian}.
In \Cref{supp:table:correspondence}, we expand on this connection and list $p(\beta)$ for  two other commonly-used priors for scale mixture of normal distributions: Jeffreys, Inverse-gamma, and Inverse-beta.
Note that the half-Cauchy distribution $p(z) \propto (1 + z^2)^{-1}$ is a special case of the inverse-beta distribution for $s = t = 1/2$.
Half-cauchy prior is another commonly used prior for compressing Bayesian neural networks~\citepapp{louizos2017bayesian}.

\begin{table}[b!]
  \begin{center}
    \begin{tabular}{c c c}
      \toprule
      Prior name & $p(z^2)$ & $p(\beta)$\\
      \midrule
      Jeffreys & $p(z^2) \propto z^{-2}$ & $p(\beta) \propto \beta^{-1}$\\
      Inverse-gamma & $p(z^2) \propto z^{-2(s+1)}e^{-t/(2z^2)}$ & $p(\beta) \propto \beta^{s-1}e^{-t\beta/2}$\\
      Inverse-beta & $p(z^2) \propto (z^2)^{t-1}(1+z^2)^{-(s+t)}$ & $p(\beta) \propto \beta^{-(s + 2t+1)} ( 1 + \beta)^{-(s+t)}$\\
      \bottomrule
    \end{tabular}
  \end{center}
  \caption{\label{supp:table:correspondence}Correspondence between $p(z^2)$ and $p(\beta)$.}
\end{table}

\subsection{MSE risk}
In this section, we prove that the Bayes estimators for the Jeffreys, inverse-gamma, and the inverse-beta (by extension the half-Cauchy) distributions dominate the maximum-likelihood estimator with respect to the mean-squared error.

Define the mean-squared error of an estimator $\hat{w}(x)$ of $\overline{w}$ as
\begin{align}
    \text{MSE}(\overline{w}, \hat{w}) = \E_{x} \|\hat{w}(x) - \overline{w}\|^2,
\end{align}
where the expectation is taken over $\mathcal{N}(x;\overline{w}, \alpha^2)$.
Immediately, we have $\text{R}(\overline{w}, \hat{w}^{(\text{ML})}) = d$, where $\hat{w}^{(\text{ML})}(x) = x$ is the maximum-likelihood estimator.
An estimator $\hat{w}^{(a)}(x)$ is said to dominate another estimator $\hat{w}^{(b)}(x)$ if $\text{MSE}(\overline{w}, \hat{w}_a) \leq \text{MSE}(\overline{w}, \hat{w}_b)$ for all $\overline{w}$ and the inequality is strict for a set of positive Lesbesgue measure.
It is well-known that the maximum-likelihood estimator is minimax~\citep{george2006improved}, and thus any estimator that dominates the maximum-likelihood estimator is also minimax.

To compute the mean-squared error risk for an estimator $\hat{w}(x)$, observe that
\begin{align}
    \|\hat{w}(x) - \overline{w}\|^2 = \|x -\hat{w}(x)\|^2 - \|x - \overline{w}\|^2 + 2 (\hat{w}(x) - \overline{w})^\top(x - \overline{w}).
\end{align}
Taking expectations on both sides gives 
\begin{align}
    \label{supp:eq:SURE}
    \text{MSE}(\overline{w}, \hat{w}) 
    &= \E_x \|x - \hat{w}(x)\|^2 - d  + 2 \sum_{i=1}^d \text{Cov}(\hat{w}_i(x), x_i)\\
    &= \E_x \|x - \hat{w}(x)\|^2 - d  + 2  \E_x \nabla \cdot \hat{w}(x)
\end{align}
where $\nabla = (\partial/\partial{x_1}, \ldots, \partial/\partial{x_d})$ and we apply Stein's lemma $\text{cov}(\hat{w}_i(x), x_i) =  \mathbb{E}_x \partial \hat{w}_i / \partial x_i$ in the last line.
If the estimator takes the form $\hat{w}(x) = x + \gamma(x)$, the expression simplifies as:
\begin{align}
    \label{supp:eq:SURE_specialized}
    \text{MSE}(\overline{w}, \hat{w}) 
    &= d + \E_x \|\gamma(x)\|^2 + 2  \E_x \nabla \cdot \gamma(x).
\end{align}
Therefore, an estimator $\hat{w}(x) = x + \gamma(x)$ dominates $\hat{w}^{(\text{ML})}(x)$ if 
\begin{align}
  \label{supp:eq:domination_condition}
  \text{MSE}(\overline{w}, \hat{w}) - \text{MSE}(\overline{w}, \hat{w}^{(\text{ML})})
    = \E_x \left [ 
      \| \gamma(x)\|^2 
      +
    2\nabla \cdot \gamma(x)
    \right ]
    \leq 0
\end{align}
for all $\overline{w}$ and the inequality is strict on a set of positive Lesbesgue measure.


\subsubsection{James-Stein estimator}
The famous Jame-Stein estimator is defined as 
\begin{align}
\hat{w}^{(\text{JS})}(x) = x + \gamma^{(\text{JS})}(x), 
\quad
\gamma^{(\text{JS})}(x) = -(d-2) x / \|x\|^2,
\end{align}
with
\begin{align}
    \nabla \cdot \gamma^{(\text{JS})}(x)
    &=
    \sum_{i=1}^d
    \left [ 
    - \frac{d-2}{\|x\|^2} 
    + 2 \frac{d-2}{(\|x\|^2)^2} x_i^2
    \right ]
    = -\frac{(d-2)^2}{\|x\|^2},\\
    \|\gamma^{(\text{JS})}(x)\|^2 
    &=
    \frac{(d-2)^2}{\|x\|^2}
    .
\end{align}
Substituting $\nabla \cdot \gamma^{(\text{JS})}(x)$ and $\|\gamma^{(\text{JS})}(x)\|^2$ into \Cref{supp:eq:domination_condition}, we have
\begin{align}
  \text{MSE}(\overline{w}, \hat{w}^{(\text{JS})}) - \text{MSE}(\overline{w}, \hat{w}^{(\text{ML})})
    &= \E_x 
      \frac{(d-2)^2}{\|x\|^2}.
\end{align}
Thus, the James-Stein estimator dominates the maximum-likelihood estimator for $d > 2$.
\subsubsection{Bayes estimators}
The Bayes estimator for a prior choice $p(\beta)$ is given by~\citepapp{brown1971admissible}:
\begin{align}
\hat{w}^{{(p(\beta))}}(x) = x + \gamma^{(p(\beta))}(x), 
\quad
\gamma^{(p(\beta))}(x) = \nabla \log m(x),
\end{align}
where
\begin{align}
m(x) &=  \int \mathcal{N}(x; 0, \beta^{-1}I_d) p(\beta) d\beta\\
    &= \int (2\pi)^{-\frac{1}2} \beta^{d/2} \exp \left ( -\beta x^2 / 2 \right ) p(\beta) d\beta.
\end{align}
Substituting $\gamma^{(p(\beta))}(x)$ into \Cref{supp:eq:domination_condition}, we find that the condition for the Bayes estimator to be minimax is given by~\citepapp{george2006improved}:
\begin{align}
  \label{supp:eq:bayes_dominate2}
\text{MSE}(\overline{w}, \hat{w}^{(\text{B})}) - \text{MSE}(\overline{w}, \hat{w}^{(\text{ML})})
    &= \E_x \left [ 
      -\|\nabla \log m(x)\|^2
      +
      2
      \frac{\nabla^2 m(x)}{m(x)}
      \right ]\\
    &=
    \E_x  \left [ 
      4 \frac{\nabla^2 \sqrt{m(x)}}{\sqrt{m(x)}}
      \right ]
      \leq 0,
\end{align}
where $\nabla^2 = \sum_i \partial^2/\partial x_i^2$ is the Laplace operator.
This condition holds when $\sqrt{m(x)}$ is superharmonic (i.e., $\sqrt{m(x)} \leq 0, \forall x \in \mathbb{R}^d$), suggesting a recipe for constructing Bayes estimators that dominate the maximum likelihood estimator, summarized in the following proposition.

\begin{restatable}[Extension of Theorem 1 in \citealp{fourdrinier1998construction}]{proposition}{bayes_prior} \label{thm:bayes_prior}
  Let $p(\beta)$ be a positive function such that $f(\beta) = \beta p'(\beta) / p(\beta)$ can be decomposed as $f_1(\beta) + f_2(\beta)$  where $f_1$ is non-decreasing, 
  $f_1 \leq A$, $0 < f_2 \leq B$, and $A / 2 + B \leq (d-6) / 4$.
  Assume also that $\lim_{\beta \to 0} \beta^{d/2+2} p(\beta) = 0$.
  Then, $\nabla^2\sqrt{m(x)} \leq 0$ and the Bayes estimator is minimax.
  If $A/2 + B < (d-6) /4$, then the Bayes estimator dominates $\hat{w}^{(\text{ML})}(x)$.
\end{restatable}

\begin{proof}
  This proof largely follows the proof of Theorem 1 in \citepapp{fourdrinier1998construction}.

  Note that \Cref{supp:eq:bayes_dominate2} holds if 
  \begin{align}
  \nabla^2 \sqrt{m(x)} = 
  \frac{1}{2 \sqrt{m(x)}}
  \left ( 
    \nabla^2 m(x) - \frac{1}{2} \frac{\|\nabla m(x)\|^2}{m(x)}
  \right )
  \leq 0
  \quad \forall x \in \mathbb{R}^d,
  \end{align}
  or equivalently
  \begin{align}
    \frac{\nabla^2 m(x)}{\|\nabla m(x)\|} - \frac{1}2 \frac{\|\nabla m(x)\|}{m(x)} \leq 0
    \quad
    \forall x \in \mathbb{R}^d.
  \end{align}
  Computing the derivatives, we get the condition
  \begin{align}
  \frac{
      \int_0^1
    \left (
      \beta \|x\|^2 - d \right ) 
      \beta^{d/2+1}
    e^{-\beta\|x\|^2/2 }
      p(\beta)
      d\beta
  }{
    \|x\|
    \int_0^1
    \beta^{d/2+1}
    e^{-\beta\|x\|^2/2 }
    p(\beta)
    d\beta
  }
  -
  \frac{1}2
  \frac{
  \|x\|
    \int_0^1
    \beta^{d/2 + 1}
    e^{-\beta\|x\|^2/2 }
    p(\beta)
    d\beta
  }{ 
  \int_0^1
    \beta^{d/2}
    e^{-\beta\|x\|^2/2 }
    p(\beta)
    d\beta
    }
  \leq 0.
  \end{align}
Divide both sides by $\|x\|$ and rearrange to get
\begin{align}
  \frac{
    \int_0^1 \beta^{d/2 + 2}
    e^{-\beta\|x\|^2/2 }
    p(\beta)
    d\beta
  }{
    \int_0^1 \beta^{d/2+1}
    e^{-\beta\|x\|^2/2 }
    p(\beta)
    d\beta
  }
  -
  \frac{1}2
  \frac{
    \int_0^1 \beta^{d/2+1}
    e^{-\beta\|x\|^2/2 }
    p(\beta)
    d\beta
  }{
\int 
    \beta^{d/2}
    e^{-\beta\|x\|^2/2 }
    p(\beta)
    d\beta
  }
  \leq \frac{d}{\|x\|^2}.
  \label{supp:eq:bayes_prior_condition_2}
\end{align}
Next, we integrate by parts the numerator of the first term on the left-hand side to get:
\begin{align}
  \int_0^1 \beta^{d/2+2}
    e^{-\beta\|x\|^2/2 }
  p(\beta) d\beta 
  &= 
  -\frac{2}{\|x\|^2} 
  \left [ 
    \beta^{d/2+2}
    e^{-\beta\|x\|^2/2 }
    p(\beta)
  \right ]_0^1\\ \nonumber
  &\hphantom{=}
  +\frac{d+4}{\|x\|^2} 
  \int_0^1 \beta^{d/2+1} 
    e^{-\beta\|x\|^2/2 }
  p(\beta) d\beta\\ \nonumber
  &\hphantom{=}
  +\frac{2}{\|x\|^2}
  \int_0^1 \beta^{d/2+2} 
    e^{-\beta\|x\|^2/2 }
  p'(\beta) d\beta,
\end{align}
where the middle term is the same as the denominator of the first term in \Cref{supp:eq:bayes_prior_condition_2}.
Integrating by parts the second term gives the same expression as that of the first term, but with $d-2$ in place of $d$ everywhere.
Substituting these expressions back into \Cref{supp:eq:bayes_prior_condition_2}, collecting like terms, and dividing both sides by $2/\|x\|^2$, gives:
\begin{align}
  &\frac{
  \int_0^1 \beta^{d/2+2} 
    e^{-\beta\|x\|^2/2 }
  p'(\beta) d\beta}
    {
    \int_0^1 \beta^{d/2+1}
    e^{-\beta\|x\|^2/2 }
    p(\beta)
    d\beta
    }
    -\frac{1}{2}
\frac{
  \int_0^1 \beta^{d/2+1} 
    e^{-\beta\|x\|^2/2 }
  p'(\beta) d\beta
    }
    {
    \int_0^1 \beta^{d/2}
    e^{-\beta\|x\|^2/2 }
    p(\beta)
    d\beta
    }
    + \kappa_0 + \kappa_1\\ \nonumber
    &\leq \frac{d}{2} - \frac{d+4}{2} + \frac{1}{2}\frac{d+2}{2}
    =\frac{d-6}{4},
\end{align}
where
\begin{align}
  \kappa_1 
  &= 
  -
  \frac{\lim_{\beta \to 1}
    \beta^{d/2+2}
    e^{-\beta\|x\|^2/2 }
    p(\beta)
  }{
    \int_0^1 \beta^{d/2+1}
    e^{-\beta\|x\|^2/2 }
    p(\beta)
    d\beta
  }
  +
  \frac{1}{2}
  \frac{\lim_{\beta \to 1}
    \beta^{d/2+1}
    e^{-\beta\|x\|^2/2 }
    p(\beta)
  }{
    \int_0^1 \beta^{d/2}
    e^{-\beta\|x\|^2/2 }
    p(\beta)
    d\beta
  },
  \\
  \kappa_0 
  &= 
  \frac{\lim_{\beta \to 0}
    \beta^{d/2+2}
    e^{-\beta\|x\|^2/2 }
    p(\beta)
  }{
    \int_0^1 \beta^{d/2+1}
    e^{-\beta\|x\|^2/2 }
    p(\beta)
    d\beta
  }
  -
  \frac{1}{2}
  \frac{\lim_{\beta \to 0}
    \beta^{d/2+1}
    e^{-\beta\|x\|^2/2 }
    p(\beta)
  }{
    \int_0^1 \beta^{d/2}
    e^{-\beta\|x\|^2/2 }
    p(\beta)
    d\beta
  }.
\end{align}
Here, both $\kappa_0$ and $\kappa_1$ are nonpositive: 
(i) $\kappa_0$ is nonpositive because the first term vanishes due to the boundary conditions and the second term is nonpositive, and
(ii) $\kappa_1$ is nonpositive because the limits of the numerators of the two terms are equal while the denominator of the second term is larger than that of the first.
We can thus drop $\kappa_0$ and $\kappa_1$ to get the sufficient condition:
\begin{align}
  \E_{d} \left ( f \right ) - \frac{1}{2} \E_{d-2} \left ( f \right ) \leq \frac{d-6}{4},
\end{align}
where $\E_d$ denotes expectation with respect to the density
\begin{equation}
  g_d(\beta) = \frac{
    \beta^{d/2+1} e^{-\beta\|x\|^2/2} p(\beta)
  }{
    \int_0^1 \beta^{d/2+1} e^{-\beta\|x\|^2/2} p(\beta) d\beta
  }
\end{equation}
and where $f(\beta) = \beta p'(\beta)/p(\beta)$.

Because $g_d(\beta)$ is a family of monotone increasing likelihood ratio in $d$ and $f_1$ is nonincreasing and bounded by $A$, we have $\E_d(f_1) - \E_{d-2}(f_1)/2 \leq A / 2$.
We have $\E_d(f_2) - \E_{d-2}(f_2) / 2 \leq B$ because $0 < f_2 \leq B$.
Taken together, we have 
\begin{align}
  \E_d(f) - \E_{d-2}(f)/2 \leq A/2 + B \leq (k-6) / 4.
\end{align}
When the inequality is strict (i.e., $A/2 + B < (k-6)/ 4$), then $\nabla^2 \sqrt{m(x)} < 0$ and the Bayes estimator dominates the maximum-likelihood estimator.
\end{proof}

Checking whether a given $p(\beta)$ satisfy the conditions in \Cref{thm:bayes_prior} may be tedious. 
The following corollary is useful for construction $p(\beta)$ that satisfies the conditions in \Cref{thm:bayes_prior}.

\begin{restatable}[Extension of Corollary 1 in \citealp{fourdrinier1998construction}]{corollary}{construction} \label{thm:construction}
  Let $\psi$ be a continuous function that can be decomposed as $\psi_1 + \psi_2$, with $\psi_1 \leq C$, $\psi_1$ non-decreasing, $0 < \psi_2 \leq D$, and $C/2 + D \leq 0$. 
  Let
  \begin{align}
    p(\beta) = \exp \left ( 
      \frac{1}2 \int_{\beta_0}^\beta
      \frac{2\psi(u) + d - 6}{u} du
      \right )
      \quad
      \forall \beta_0 \geq 0,
  \end{align}
  such that $\lim_{\beta \to 0} \beta^{d/2 + 2} p(\beta) = 0$ and $\beta_0 \in (0, 1)$ is a constant.
  Then, $p(\beta)$ results in a minimax Bayes estimator, which  dominates the maximum likelihood estimator when $C / 2 + D < 0$.
\end{restatable}
\begin{proof}
  The proof is the same as that of Corollary 1 in \citealpapp{fourdrinier1998construction}, with \Cref{thm:bayes_prior} in place of Theorem 1 in \citealpapp{fourdrinier1998construction}. 
\end{proof}

Using \Cref{thm:construction}, we now check that the three priors listed in \Cref{supp:table:correspondence} and referenced in \Cref{subsec:sparsity} lead to Bayes estimators that dominate the maximum-likelihood estimator.

\paragraph{Jeffreys prior} %
Let $\psi_1(u) = a$ for $a \leq 0$ and $\psi_2(u) = 0$.
We have
\begin{align}
  p(\beta) = 
  \exp \left ( 
    \frac{1}2
  \int_{\beta_0}^\beta 
  \frac{2 a + d-6}{u}
  du
  \right ) 
          \propto \beta^{a + (d-6)/ 2}.
\end{align}
To satisfy $\lim_{\beta \to 0} \beta^{d/2+2}p(\beta) = 0$, we require $1 -d < a \leq 0$.
We recover the improper normal-Jeffreys prior $p(\beta) \propto \beta^{-1}$, for $a = 2 - d/2$.
The corresponding Bayes estimator dominates the maximum likelihood estimator when $d > 4$.

\paragraph{Inverse-gamma prior}
Let $\psi_1(u) = a$ and $\psi_2(u) = b (1-u)/ 2$ for $a\leq 0$ and $b\geq 0$. 
We have
\begin{align}
  p(\beta) = 
  \exp \left ( 
  \int_{\beta_0}^\beta 
  \frac{a + b(1-u)/2 + (d-6) / 2}{u}
  du
  \right ) 
          \propto 
          \beta^{a + (b + d-6)/2} e^{-b\beta/2}.
\end{align}
Setting $C=a$ and $D=b/2$, we get the followings conditions: $a+b \leq 0$ and $1-d \leq a + b/2$.
Note that when these conditions are met with $s=a+(b+d-4)/2$ and $t=b$, we recover the inverse-gamma prior in \Cref{supp:table:correspondence}.

\paragraph{Inverse-beta (half-Cauchy) prior}
Let $\psi_1(u) = a$ and $\psi_2(u) = b / (u+1)$ for $a\leq 0$ and $b \geq 0$. 
We have
\begin{align}
  p(\beta) = 
  \exp \left ( 
  \int_{\beta_0}^\beta 
  \frac{a + b/ (1+u) + (d-6) / 2}{u}
  du
  \right ) 
          \propto 
          \beta^{a + b + (d-6)/2} (1 + \beta)^{-b}.
\end{align}
Setting $C=a$ and $D=b$, we get the condition $a/2 + b \leq 0$.
To satisfy $\lim_{\beta \to 0} \beta^{d/2+2}p(\beta) = 0$, we require $1 -d < a + b \leq 0$.
Note that this corresponds to the inverse-beta prior in \Cref{supp:table:correspondence} with $t=a+(d-8)/2$ and $s=b-t$.

To recover the half-Cauchy prior, we set $b=1$ and $a=(5-d)/2$.
All conditions in \Cref{thm:construction} are satisfied when $d>9$.

\section{Limitations} \label{sect:limits}
One weakness of the current theoretical analysis regarding the choice of sparsity-inducing priors is the assumption of Gaussian (and in particular, isotropic Gaussian) structure in the parameter space of optimal policies for clusters of tasks. In reality, there is likely a nontrivial degree of covariance among task parameterizations. Extending our analysis to more realistic forms of task structure is an important direction for future work. In a similar vein, the assumption that tasks are drawn iid from a fixed distribution is also unrealistic in naturalistic settings. It would be interesting to introduce some form of sequential structure (e.g., tasks are drawn from a Markov process). Another direction for future work is expanding beyond the ``one control policy, one default policy'' setup--having, for example, one default policy per task cluster and the ability to reuse and select (for example, using successor feature-like representations \citep{Barreto:2020_fastgpi,Barth:2018,moskovitz2022_fr}) among an actively-maintained set of control policies across tasks and task clusters would be useful.

\section{OCO Background} \label{sect:oco}
In \textit{online convex optimization} (OCO), the learner observes a series of convex loss functions $\ell_k: \mathsf N \to \reals$, $k=1,\dots,K$, where $\mathsf N \subseteq \reals^d$ is a convex set. After each round, the learner produces an output $x_k \in \mathsf N$ for which it will then incur a loss $\ell_k(x_k)$ \citep{orabona_online19}. At round $k$, the learner is usually assumed to have knowledge of $\ell_1, \dots, \ell_{k-1}$, but no other assumptions are made about the sequence of loss functions. The learner's goal is to minimize its average regret:
\begin{align}
  \bar{\mathcal R}_K \coloneqq \frac{1}{K}\sum_{k=1}^K \ell_k(x_k) - \min_{x\in\mathsf N} \frac{1}{K}\sum_{k=1}^K \ell_k(x). 
\end{align}
One OCO algorithm which enjoys sublinear regret is \textit{follow the regularized leader} (FTRL). In each round of FTRL, the learner selects the solution $x\in\mathsf N$ according to the following objective:
\begin{align}
  x_{k+1} = \argmin_{x\in\mathsf N} \psi_k(x) + \sum_{i=1}^{k-1}\ell_i(x),
\end{align}
where $\psi_k: \mathsf N\to\reals$ is a convex regularization function.

\section{Proofs of Performance Bounds and Additional Theoretical Results} \label{sect:performance_proofs}

The following result is useful. 
\begin{lemma} \label{thm:lipschitz}
The function $\ell(\nu) = \E_{w\sim\nu}f(w)$ is $L$-Lipschitz with respect to the TV distance as long as $f: \mathcal W \to \reals$ lies within $[0, L]$  $\forall w\in\mathcal W$, $\mathcal W\subseteq \reals^d$ for some $L<\infty$.  
\end{lemma}

\begin{proof}
  We have 
  \begin{align*}
    |\ell(\nu_1) - \ell(\nu_2)| &= \left\vert\E_{w\sim\nu_1} f(w) - \E_{w\sim\nu_2}f(w)\right\vert \\ 
      &= \left\vert\int_{\mathcal W} (\nu_1(w) - \nu_2(w)) f(w)\ dw\right\vert \\ 
      &\leq \sup_{w\in\mathcal W} |(\nu_1(w) - \nu_2(w)) f(w)| \\
      &\leq L \sup_{w\in\mathcal W} |\nu_1(w) - \nu_2(w)| \\
      &= L d_{\mathrm{TV}}(\nu_1, \nu_2).
  \end{align*}
\end{proof}

\begin{restatable}[Default Policy Distribution Regret]{proposition}{defaultregret} \label{thm:default_regret}
  Let tasks $M_k$ be independently drawn from $\pr_\mathcal M$ at every round, and let them each be associated with a deterministic optimal policy $\pi_k^\star: \St \to \A$.  
  We make the following mild assumptions: i) $\pi_w(a^\star|s) \geq \epsilon > 0$ $\forall s\in\St$, where $a^\star=\pi_k^\star(s)$ and $\epsilon$ is a constant. ii) $\min_\nu \kl[\nu(\cdot), p(\cdot)] \to 0$ as $\mathrm{Var}[\nu] \to \infty$ for an appropriate choice of sparsity-inducing prior $p$. 
  Then \cref{alg:mdlc} guarantees 
  \begin{align} \label{eq:ub_pi0}
    \E_{\pr_\mathcal M} [\ell_K(\nu_K) - \ell_K(\bar\nu_K)] \leq \left(\E_{\pr_\mathcal M}\kl[\bar\nu_K, p] + 1\right) \frac{\log (1/\epsilon)}{\sqrt{K}}. 
  \end{align}
  where $\bar\nu_K = \argmin_{\nu\in\mathsf N} \sum_{k=1}^K \ell_k(\nu)$. 
\end{restatable}

\begin{proof}
  The first part of the proof sets up an application of \citeapp{orabona_online19}, Corollary 7.9. 
  
  To establish grounds for its application, we first note the standard result that the regularization functional $\psi(\nu) = \kl[\nu(w), p(w)]$ for probability measures $\nu, p \in \mathcal P(\mathcal W)$ is $1$-strongly convex in $\nu$ \citep{melbourne_2020kl}. 
  
  Finally, assumption (i) implies that the KL between the default policy and the optimal policy is upper-bounded: $\kl[\pi_{k}^\star, \pi_w] \leq \log 1/\epsilon$. Then by \cref{thm:lipschitz}, $\ell_k(\nu)$ is $L$-Lipschitz wrt the TV distance, where $L = \log1/\epsilon$. 

  Note also that under a Gaussian parameterization for $\nu$, the distribution space $\mathsf N$ is the Gaussian parameter space $\mathsf N = \{(\mu, \Sigma) \ : \ \mu\in\reals^d,\  \Sigma\in\reals^{d\times d}, \Sigma \succeq 0\}$, which is convex \citep{boyd:optimization_06}. 
  
  Then \citeapp{orabona_online19}, Corollary 7.9 gives 
  \begin{align} \label{eq:ftrl}
      \frac{1}{K} \sum_{k=1}^K \ell_k(\nu_k)  - \frac{1}{K}\sum_{k=1}^K \ell_k(\bar\nu_K) \leq \left(\frac{1}{\alpha}\kl[\bar\nu_K, p] + \alpha \right)\frac{L}{\sqrt{K}},
  \end{align}
  where $\bar\nu_K = \argmin_\nu \sum_{k=1}^K \ell_k(\nu)$. The constant $\alpha\in\reals^+$ is a hyperparameter, so we are free to set it to $1$ \citep{orabona_online19}. Finally, we observe that $\E_{\pr_{\mathcal M_i}} \frac{1}{K}\sum_{k=1}^K \ell(\nu_k) = \E_{\pr_{\mathcal M_i}}\ell_K(\nu_K)$ and take the expectation with respect to $\pr_{\mathcal M_i}$ of both sides of \cref{eq:ftrl} to get the desired result: 
  \begin{align}
      \E_{\pr_{\mathcal M_i}} [\ell_K(\nu_K) - \ell_K(\bar\nu_{K})]
      \leq \left(\E_{\pr_{\mathcal M_i}}\kl[\bar\nu_K, p] + 1 \right)\frac{L}{\sqrt{K}}.
  \end{align}
\end{proof}

\controlperf*

Note: In the above, there is a small error---it should be $\alpha_k(s) \coloneqq \E_{w\sim\nu}\tv(\pi_k^\star(\cdot|s), \pi_w(\cdot|s)) \leq \sqrt{\frac{1}{2}G(K)}$. $d_\rho^\pi$ refers to the discounted state-occupancy distribution under $\pi$ with initial state distribution $\rho$:
\begin{align}
  d_\rho^\pi(s) = \E_{s_0\sim\rho} (1 - \gamma) \sum_{h\geq 0} \gamma^h \pr^\pi(s_h=s|s_0). 
\end{align}
Division between probability mass functions is assumed to be element-wise.

\begin{proof}
  Without loss of generality, we prove the bound for a fixed state $s\in\St$, noting that the bound applies independently of our choice of $s$. We use the shorthand $\kl[\pi(\cdot|s), \pi_w(\cdot|s)] \to \kl[\pi, \pi_w]$ for brevity. We start by multiplying both sides of the bound from \cref{thm:default_regret} by $1/2$ and rearranging:
  \begin{align}
  \begin{split}
      &\hphantom{\geq}
      \frac{1}{2}
      \left (
      \E_{\pr_{\mathcal M_i}} \ell_K(\bar\nu_K) + 
      \frac{L}{\sqrt{K}}
      \left(\E_{\pr_{\mathcal M_i}}\kl[\bar\nu_K, p] + 1 \right) 
      \right )
      \\
      &\geq \E_{\pr_{\mathcal M_i}} \frac{1}{2}\ell_K(\nu_K) \\
      &= \E_{\pr_{\mathcal M_i}}\E_{\nu_K} \frac{1}{2}\kl[\pi_K^\star, \pi_w] \\
      &\stackrel{(i)}{=} \E_{\pr_{\mathcal M_i}}\left[ \mathrm{Var}_{\nu_K}\left[ \sqrt{\frac{1}{2}\kl[\pi_K^\star, \pi_w]}\right] + \E_{\nu_K}\left[\sqrt{\frac{1}{2}\kl[\pi_K^\star, \pi_w]}\right]^2\right] \\
      &\stackrel{(ii)}{\geq} \E_{\pr_{\mathcal M_i}}\left[  \E_{\nu_K}\left[\sqrt{\frac{1}{2}\kl[\pi_K^\star, \pi_w]}\right]^2\right]
  \end{split}
  \end{align}
  where $(i)$ follows from the definition of the variance, and $(ii)$ follows from its non-negativity. 
  We can rearrange to get
  \begin{align}
  \begin{split}
  \frac{L}{2\sqrt{K}}
  \left(\E_{\pr_{\mathcal M_i}}\kl[\bar\nu_K, p] + 1 \right)
       &\geq 
      \E_{\pr_{\mathcal M_i}}\E_{\nu_K}\left[\sqrt{\frac{1}{2}\kl[\pi_K^\star, \pi_w]}\right]^2\\
      &\stackrel{(ii)}{\geq} \E_{\pr_{\mathcal M_i}}\E_{\nu_K}\left[d_{\mathrm{TV}}(\pi_K^\star, \pi_w)\right]^2
  \end{split}
  \end{align}
  where $(ii)$ follows from Pinsker's inequality. Letting $\alpha_K(s) = \sqrt{\frac{1}{2}G(K)}$ and applying \cite{Moskovitz:2021_rpotheory}, Lemma 5.2 gives the desired result. 
\end{proof}

This upper-bound is signficant, as it shows that, all else being equal, a high complexity barycenter default policy distribution $\bar\nu_K$ (where complexity is measured by $\kl[\bar\nu_K, p]$) leads to a slower convergence rate in the control policy.

\begin{algorithm}[!t]
	\caption{Idealized MDL-C for Multitask Learning}\label{alg:mdlc}
		\begin{algorithmic}[1] 
		    \STATE require: task distribution $\pr_\mathcal M$, policy class $\Pi$, coefficients $\{\eta_{k}\}$
		    \STATE initialize: default policy distribution  $\nu_1\in\mathsf{N}$
			\FOR{tasks $k=1, 2, \dots, K $}
			    \STATE Sample a task $M_k\sim \mathcal P_\mathcal M(\cdot)$
                \STATE Optimize control policy: 
                \begin{equation} 
                    \hat\pi_k^\star = \argmax_{\pi\in\Pi} V^\pi_{M_k} - \lambda \E_{s\sim d^\pi}\E_{w\sim\nu_k}\kl[\pi_w(a|s), \pi(a|s)]
						\label{algo:eq:control}
                \end{equation}
			    \STATE Update default policy distribution:
                    \begin{align}
                        \nu_{k+1} = \argmin_{\nu \in \mathsf N} \kl[\nu, p] + \E_{w\sim\nu}\kl[\hat\pi_k^\star, \pi_w]
						\label{algo:eq:default}
                    \end{align}
			 \ENDFOR
	\end{algorithmic}
\end{algorithm}

\subsection{MDL-C with Persistent Replay}
Rather than rely on iid task draws to yield a bound on the expected regret under the task distribution, a more general formulation of MDL-C for sequential task learning is described in \cref{alg:mdlc_replay}. In this setting, the dataset of optimal agent-environment interactions is explicitly constructed by way of a replay buffer which persists across tasks and is used to train the default policy distribution. This is much more directly in line with standard FTRL, and we can obtain the standard FTRL bound. 

\begin{restatable}[Persistent Replay FTRL Regret; \citep{orabona_online19}, Corollary 7.9]{proposition}{persistentregret} \label{thm:persistent_regret}
  Let tasks $M_k$ be independently drawn from $\pr_\mathcal M$ at every round, and let them each be associated with a deterministic optimal policy $\pi_k^\star: \St \to \A$.  
  We make the following mild assumptions: i) $\pi_w(a^\star|s) \geq \epsilon > 0$ $\forall s\in\St$, where $a^\star=\pi_k^\star(s)$ and $\epsilon$ is a constant. ii) $\min_\nu \kl[\nu(\cdot), p(\cdot)] = 0$ asymptotically as $\mathrm{Var}[\nu] \to \infty$. 
  Then with $\eta_{k-1} = L\sqrt{k}$, \cref{alg:mdlc_replay} guarantees 
  \begin{align} \label{eq:persistent_pi0_ub}
    \frac{1}{K} \sum_{k=1}^K \ell_k(\nu_k)  - \frac{1}{K}\sum_{k=1}^K \ell_k(\bar\nu_K) \leq \left(\kl[\bar\nu_K, p] + 1 \right)\frac{L}{\sqrt{K}},
\end{align}
  where $\bar\nu_K = \argmin_{\nu\in\mathsf N} \sum_{k=1}^K \ell_k(\nu)$. 
\end{restatable}
\begin{proof}
  This follows directly from the arguments made in the proof of \cref{thm:default_regret}. 
\end{proof}

As before, this result can be used to obtain a performance bound for the control policy. 

\begin{restatable}[Control Policy Sample Complexity for MDL-C with Persistent Replay]{proposition}{controlperf_replay} \label{thm:controlperf_replay}
  Under the setting described in \cref{thm:persistent_regret}, denote by $T_k$ the number of iterations to reach $\epsilon$-error for $M_k$ in the sense that $
        \min_{t\leq T_k} \{V^{\pi_k^\star} - V^{(t)}\} \leq \eps
    $
    and the upper-bound in \cref{eq:persistent_pi0_ub} by $G(K)$. In a finite MDP, from any initial $\theta^{(0)}$, and following gradient ascent, $\expect[M_k\sim\mathcal P_\mathcal M]{T_k}$ satisfies:
    \begin{align*}
        \expect[M_k\sim\mathcal P_{\mathcal M_i}]{T_k} \geq  \frac{80\vert\mathcal{A} \vert^2 \vert \mathcal{S}\vert^2}{\epsilon^2(1-\gamma)^6} \mathbb E_{\substack{M_k\sim\mathcal P_{\mathcal M_i}
        s\sim\Unif_\St}}\left[\kappa_\A^{\alpha_k}(s) \left\| \frac{d_{\rho}^{\pi_k^*} }{\mu}  \right \|_\infty^2\right],
    \end{align*}
    where $\alpha_k(s) \coloneqq \E_{w\sim\nu}\tv(\pi_k^\star(\cdot|s), \pi_w(\cdot|s)) \leq \sqrt{\frac{1}{2}G(K)}$, $\kappa_\A^{\alpha_k}(s) = \frac{2|\A|(1 - \alpha(s)) }{2|\A|(1 - \alpha(s)) - 1}$, and $\mu$ is a measure over $\St$ such that $\mu(s)>0$ $\forall s\in\St$. 
\end{restatable}

\begin{proof}
  Without loss of generalization we select a single state $s\in\St$, observing that the same analysis applies $\forall s\in\St$. For simplicity, we denote $\pi(\cdot|s) = \pi$. We start by multiplying each side of \cref{eq:ftrl} by $K$ and rearranging:
  \begin{align}
  \begin{split}
    \sum_{k=1}^K \ell_k(\nu_k) - \sum_{k=1}^K \ell_k(\bar\nu_K) &\leq \left(\kl[\bar\nu, p] + 1\right)L\sqrt{K} \\
    \Rightarrow \ell_K(\nu_K) &\leq \sum_{k=1}^K \ell_k(\bar\nu_K) - \sum_{k=1}^{K-1} \ell_k(\nu_k) + \left(\kl[\bar\nu, p] + 1\right)L\sqrt{K} \\
    &= \underbrace{\ell_K(\bar\nu_K) + \sum_{k=1}^{K-1} (\ell_k(\bar\nu_K) - \ell_k(\nu_k)) + \left(\kl[\bar\nu, p] + 1\right)L\sqrt{K}}_{\coloneqq G(K)}
  \end{split}
  \end{align}
  We can multiply both sides by $1/2$ and expand $\ell_K(\nu_K)$:
  \begin{align}
  \begin{split}
    \frac{1}{2}G(K) &\geq \E_{w\sim\nu_K} \frac{1}{2} \kl[\pi_K^\star, \pi_w] \\ 
    &\stackrel{(i)}{=}  \mathrm{Var}_{\nu_K}\left[ \sqrt{\frac{1}{2}\kl[\pi_K^\star, \pi_w]}\right] + \E_{\nu_K}\left[\sqrt{\frac{1}{2}\kl[\pi_K^\star, \pi_w]}\right]^2 \\
    &\stackrel{(ii)}{\geq}  \left(\E_{\nu_K}\left[\sqrt{\frac{1}{2}\kl[\pi_K^\star, \pi_w]}\right]\right)^2 \\
    &\stackrel{(iii)}{\geq} \left(\E_{\nu_K} d_{\mathrm{TV}}(\pi_K^\star, \pi_w)\right)^2
  \end{split}
  \end{align} 
  where $(i)$ follows from the definition of the variance, $(ii)$ follows from its non-negativity, and $(iii)$ follows from Pinsker's inequality. We then have 
  \begin{align}
    \E_{\nu_K} d_{\mathrm{TV}}(\pi_K^\star, \pi_w) \leq \sqrt{\frac{1}{2}G(K)}.
  \end{align}
  Letting $\alpha_K(s) = \sqrt{\frac{1}{2}G(K)}$ and applying \citeapp{Moskovitz:2021_rpotheory}, Lemma 5.2 gives the desired result. 
\end{proof}

\subsection{Parallel Task Setting}
\begin{algorithm}[!t]
	\caption{Off-Policy MDL-C for Parallel Multitask Learning }\label{alg:mdlc_parallel}
		\begin{algorithmic}[1] 
		    \STATE require: task distribution $\pr_\mathcal M$, policy class $\Pi$
		    \STATE initialize: default policy distribution  $\nu_1\in\mathsf{N}$, control replay $\mathcal D_0 \gets \emptyset$, default replay $\mathcal D_0^\phi \gets \emptyset$
        \STATE initialize control policy parameters $\theta$ and default policy distribution parameters $\phi$. 
      \WHILE{not done}
			\FOR{episodes $k=1, 2, \dots, K $}
			    \STATE Sample a task $M_k\sim \mathcal P_\mathcal M(\cdot)$ with goal ID feature $g_k$ 
          \STATE Collect trajectory $\tau = (\tilde s_0, a_0, r_0, \dots, \tilde s_{H-1}, a_{H-1}, r_{H-1})\sim \pr^{\pi_\theta}(\cdot)$, store experience 
            \begin{align}
              \mathcal D_k \gets \mathcal D_{k-1} \cup \{(\tilde s_h, a_h, r_h, \tilde s_{h+1})\}_{h=0}^{H-1} 
            \end{align}
            where $\tilde s_h \coloneqq (s_h, g_k)$. 
          \IF{$R(\tau) \geq R^\star$ (i.e., $\pi_\theta \approx \pi^\star_k$)}
            \STATE Add to default policy replay:
              \begin{align}
                \mathcal D_k^\phi \gets \mathcal D_{k-1}^\phi \cup \{(\tilde s_h, \pi_\theta(\cdot|\tilde s_h)\}_{h=0}^{H-1}
              \end{align}
              Note that, e.g., when $\pi_\theta(a|\tilde s) = \mathcal N(a; \mu(\tilde s, g_k), \Sigma(\tilde s, g_k))$ is a Gaussian policy, $\mu(\tilde s_h, g_k), \Sigma(\tilde s_h, g_k)$ are added to the replay with $\tilde s_h$. 
          \ENDIF 
        \ENDFOR
          \STATE Update $Q$-function(s) as in \cite{haarnoja_2018sac}. 
          \STATE Update control policy: 
          \begin{equation} 
              \theta \gets \argmin_{\theta'} \E_{\mathrm{Unif}_{\mathcal D_k}}\left[V^{\pi_{\theta'}} - \alpha \E_{w\sim\nu_\phi} \kl[\pi_{\theta'}(\cdot|\tilde s_h), \pi_w(\cdot|\tilde s_h)]\right]
          \end{equation}
          \STATE Update default policy distribution: 
            \begin{align}
              \phi \gets \argmin_{\phi'} \kl[\nu_{\phi'}(\cdot), p(\cdot)] + \E_{\mathrm{Unif}_{\mathcal D_k^\phi}}\E_{w\sim\nu}\kl[\pi_\theta(\cdot|\tilde s_h), \pi_w(\cdot|\tilde s_h)]
            \end{align}
       \ENDWHILE
	\end{algorithmic}
\end{algorithm}
An overview of MDL-C as applied in the parallel task setting is presented in \cref{alg:mdlc_parallel}. One important feature to note is the return threshold $R^\star$. As a proxy for the control policy converging to $\pi_k^\star$, data are only added to the default policy replay buffer when a trajectory return is above this threshold performance (on DM control suite tasks, $R^\star$ corresponded to a test reward of at least 700). We leave more in-depth theoretical analysis of this setting to future work, but note that as the task experience is interleaved, $\bar\pi_w = \E_\nu \pi_w$ will converge to the prior-weighted KL barycenter. If, in expectation, this distribution is a TV distance of less than $1 - 1/|\A|$ from $\pi_k^\star$, then the control policy will converge faster than for log-barrier regularization \citep{Moskovitz:2021_rpotheory}.

\section{Additional Experimental Details} \label{sect:expt_details}
Below, we describe experimental details for the two environment domains in the paper. 

\subsection{FourRooms}
\paragraph{Environment} The \textsc{FourRooms} experiments are set in an $11\times11$ gridworld. The actions available to the agent are the four cardinal directions, \texttt{up}, \texttt{down}, \texttt{left}, and \texttt{right}, and transitions are deterministic. In both \textsc{FourRooms} experiments, the agent can begin an episode anywhere in the environment (sampled uniformly at random), and a single location with reward $r=50$ is sampled at the beginning of each episode from a set of possible goal states which varies depending on the experiment and the current phase. A reward of $r=-1$ is given if the agent contacts the walls. All other states give a reward of zero. Episodes end when either a time (number of timesteps) limit is reached or the agent reaches the goal state. Observations were 16-dimensional vectors consisting of the current state index (1d), flattened $3\times3$ local window surrounding the agent (includes walls, but not goals), a one-hot encoding of the action on the previous timestep (4d), the reward on the previous timestep (1d), and the state index of the current goal (1d). In the ``goal generalization'' experiment, goals may be sampled anywhere in either the top left or bottom right rooms in the first phase and either the top right or bottom left rooms in the second phase. Each phase consistent of 20,000 episodes. In the first phase, the agent was allowed 100 steps per episode, and in the second phase 25 steps.  In the ``contingency change'' experiment, the possible reward states in each phase were the top left state and bottom right state. In the second phase of training, however, the semantics of the goal feature change from indicating the location of the reward to the location where it is absent. Each phase consisted of 8,000 episodes with maximum length 100 timesteps. Results are averaged over 10 random seeds. 

\paragraph{Agents} All agents were trained on-policy with advantage actor-critic \citep{Mnih:2016_a3c}. The architecture was a single-layer LSTM \citep{hochreiter1997long} with 128 hidden units. To produce the feature sensitivity plots in \cref{fig:4rooms}c, a gating function was added to the input layer of the network:  
\begin{align}
  x_h = \sigma(b\kappa) \odot o_h,
\end{align}
where $o_h$ is the current observation, $\sigma(\cdot)$ was the sigmoid funcion, $b\in\reals$ is a constant (set to $b=150$ in all experiments), $x_h\in\reals^d$ is the filter layer output, and $\kappa\in\reals^d$ is a parameter trained using backpropagation. In this way, as $\kappa_d\to\infty$, $\sigma(b\kappa_d) \to1$, allowing input feature $o_h,d$ through the gate. As $\kappa_d\to-\infty$, the gate is shut. The plots in \cref{fig:4rooms}c track $\sigma(b\kappa_d)$ over the course of training. The baseline agent objective functions are as follows:
\begin{align}
\begin{split} \label{eq:baseline_objs}
  \mathcal J^{\mathrm{PO}}(\theta) &= V^{\pi_\theta} + \alpha \E_{s\sim d^{\pi_\theta}} \mathsf H[\pi_\theta(\cdot|s)] \\
  \mathcal J^{\mathrm{RPO}}(\theta, \phi) &= V^{\pi_\theta} - \alpha \E_{s\sim d^{\pi_\theta}} \kl[\pi_\theta(\cdot|s), \pi_\phi(\cdot|s)] \\
  \mathcal J^{\mathrm{VDO-PO}}(\theta) &= \E_{w\sim\nu_\theta}V^{\pi_w} - \beta \kl[\nu_\theta(\cdot), p(\cdot)] \\
  \mathcal J^{\mathrm{ManualIA}}(\theta, \phi) &= V^{\pi_\theta} - \alpha \E_{s\sim d^{\pi_\theta}} \kl[\pi_\theta(\cdot|s), \pi_\phi(\cdot|s_d)]; \quad s_d = s \setminus g.
\end{split}
\end{align}
In all cases $\alpha = 0.1$, $\beta=1.0$, and learning rates for all agents were set to $0.0007$. Agents were optimized with Adam \citep{kingma_ba:adam_14}. Agent control policies were reset after phase 1. 

\subsection{DeepMind Control Suite}

\paragraph{Environments/Task Settings}
We use the \texttt{walker} and \texttt{cartpole} environments from the DeepMind Control Suite \citepapp{tassa_2018DMC}. We consider two multitask settings: sequential tasks and parallel tasks. All results are averaged over 10 random seeds, and agents are trained for 500k timesteps. In the sequential task setting, tasks are sampled one at a time without replacement and solved by the agent. The control policy is reset after each task, but the default policy is preserved. For methods which have a default policy which can be preserved, performance on task $k$ is averaged over runs with all possible previous tasks in all possible orders. For example, when \texttt{walker-run} is the third task, performance is averaged over previous tasks being \texttt{stand} then \texttt{walk} and \texttt{walk} then \texttt{stand}. In the parallel task setting, a different task is sampled randomly at the start of each episode, and a one-hot task ID vector is appended to the state observation. Learning was done directly from states, not from pixels. 

\paragraph{Agents}
The base agent in all cases was SAC with automatic temperature tuning, following \citeapp{haarnoja_2018sac}. Standard SAC seeks to optimize the maximum-entropy RL objective:
\begin{align}
  \mathcal J^{\mathrm{max-ent}}(\pi) = V^\pi + \alpha\E_{s\sim d^\pi}\mathsf H[\pi(\cdot|s)] = V^\pi + \alpha\E_{s\sim d^\pi}\kl[\pi(\cdot|s), \mathrm{Unif}_\A]
\end{align}
Effectively, then, SAC uses a uniform default policy. The RPO algorithms with learned default policies replace $\kl[\pi(\cdot|s), \mathrm{Unif}_\A]$ with  $\kl[\pi(\cdot|s), \pi_w(\cdot|s)]$ (or $\kl[\pi_w(\cdot|s), \pi(\cdot|s)]$). As MDL-C requires that the control policy approximate the optimal policy before being used to generated the a learning signal for the default policy, in the sequential setting, the default policy is updated only after halfway through training. Because variational dropout can cause the network to over-sparsify (and not learn the learn adequately) if turned on too early in training, we follow the strategy of \citeapp{molchanov2017variational}, linearly ramping up a coefficient $\beta$ on the variational dropout KL from 0 to 1 starting from 70\% through training to 80\% through training. Note that \textsc{ManualIA} is not applicable to the sequential task setting, as there is no explicit goal feature. In the parallel task setting, we convert the base SAC agent into the ``multitask'' variant used by \citeapp{yu_metaworld19}, in which the agent learns a vector of temperature parameters $[\alpha_1, \dots, \alpha_K]$, one for each task. Test performance was computed by averaging performance across all $K$ tasks presented to the agent. The baseline agent objectives are as in \cref{eq:baseline_objs}. Hyperparameters shared by all agents can be viewed in \cref{table:dmc_hyperparams}.

\begin{table}[ht]
  \begin{center}
  \begin{tabular}{lr}
  \toprule
   \textbf{Hyperparameter}    &    \textbf{Value}  \\
  \midrule
  Collection Steps        &    1000  \\
   Random Action Steps       &  10000  \\
   Network Hidden Layers               &   256:256 \\
   Learning Rate          &     $3 \times 10^{-4}$ \\
   Optimizer &   Adam \\
   Replay Buffer Size      &  $1 \times 10^6$  \\
   Action Limit    & $[-1, 1]$ \\
   Exponential Moving Avg. Parameters &  $5 \times 10^{-3}$ \\
   (Critic Update:Environment Step) Ratio &  1 \\
   (Policy Update:Environment Step) Ratio & 1 \\
   Expected KL/Entropy Target &  $-\mathrm{dim}(\mathcal A)^\ast$ \\ 
   Policy Log-Variance Limits &  $[-20,2]$ \\ 
  \hline
  \end{tabular}
  \end{center}
  \caption{DM control suite hyperparameters, used for all experiments. $^\ast$The target was set to 0 for methods with learned default policies.}
  \label{table:dmc_hyperparams}
  \end{table}

\section{Additional Experimental Results} \label{sect:add_figs}

\subsection{FourRooms}

\begin{table}[ht] 
\begin{center}
  \begin{tabular}{lrr} 
  \toprule
  \textbf{Method}
                                 & \textbf{Goal Change}                              & \textbf{Contingency Change}                       \\ \midrule
  PO                             & 1.25e5 $\pm$ 1.76e4                      & 8.80e4 $\pm$ 1.64e4                      \\ 
   {RPO}      &  {1.77e5 $\pm$ 1.11e4} &  {1.04e5 $\pm$ 2.20e4} \\ 
   {VDO-PO}   &  {1.48e5 $\pm$ 1.91e4} &  {8.23e4 $\pm$ 1.98e4} \\ 
   {ManualIA} &  {1.23e5 $\pm$ 2.51e4} &  {7.69e4 $\pm$ 2.89e4} \\ 
  MDL-C                          & 1.08e5 $\pm$ 2.44e4                      & 5.11e4 $\pm$ 1.70e4                      \\ \hline
  \end{tabular}
  \end{center}
  \label{table:4rooms}
  \caption{FourRooms: Average cumulative regret across 8 random seeds in phase 2 of the goal change and contingency change experiments for each method. $\pm$ values are standard error.}
  \end{table}

\begin{table}[ht]
\begin{center}
  \begin{tabular}{lrr} 
  \toprule
  \textbf{Method} & \textbf{Cartpole}  & \textbf{Walker}                       \\ \midrule
  SAC  & 1.25e5 $\pm$ 1.76e  & 3.42e5 $\pm$ 6.10e4  \\ 
   RPO-SAC ($k=3$)  &  {1.77e5 $\pm$ 1.11e4} &  {1.04e5 $\pm$ 2.20e4} \\ 
   VDO-SAC   &  {1.48e5 $\pm$ 1.91e4} &  {8.23e4 $\pm$ 1.98e4} \\ 
   MDL-C ($k=1$) &  {1.23e5 $\pm$ 2.51e4} &  {7.69e4 $\pm$ 2.89e4} \\ 
  MDL-C  ($k=2$)                       & 1.08e5 $\pm$ 2.44e4                      & 5.11e4 $\pm$ 1.70e4                      \\ 
  MDL-C  ($k=3$)                       & 1.08e5 $\pm$ 2.44e4                      & 5.11e4 $\pm$ 1.70e4                      \\ \hline
  \end{tabular}
  \end{center}
  \label{table:dmc_seq}
  \caption{DM Control Suite, Sequential: Average cumulative regret across 8 random seeds in the sequential setting. $\pm$ values are standard error.}
  \end{table}

\begin{table}[ht]
\begin{center}
  \begin{tabular}{lrr} 
  \toprule
  \textbf{Method} & \textbf{Cartpole}  & \textbf{Walker}    \\ \midrule
  SAC  & 1.01e5 $\pm$ 2.01e3  & 1.46e5 $\pm$ 5.11e3  \\ 
   ManualIA     &  {9.90e4 $\pm$ 1.87e3} &  {1.50e5 $\pm$ 3.86e3} \\ 
  MDL-C & 9.47e4 $\pm$ 8.36e2   & 1.31e5 $\pm$ 1.35e3      \\ \hline
  \end{tabular}
  \end{center}
  \label{table:dmc_para}
  \caption{DM Control Suite, Parallel: Average cumulative regret across 8 random seeds in the parallel task setting. $\pm$ values are standard error.}
  \end{table}

\begin{figure*}[ht]
  \centering
  \includegraphics[width=0.5\textwidth]{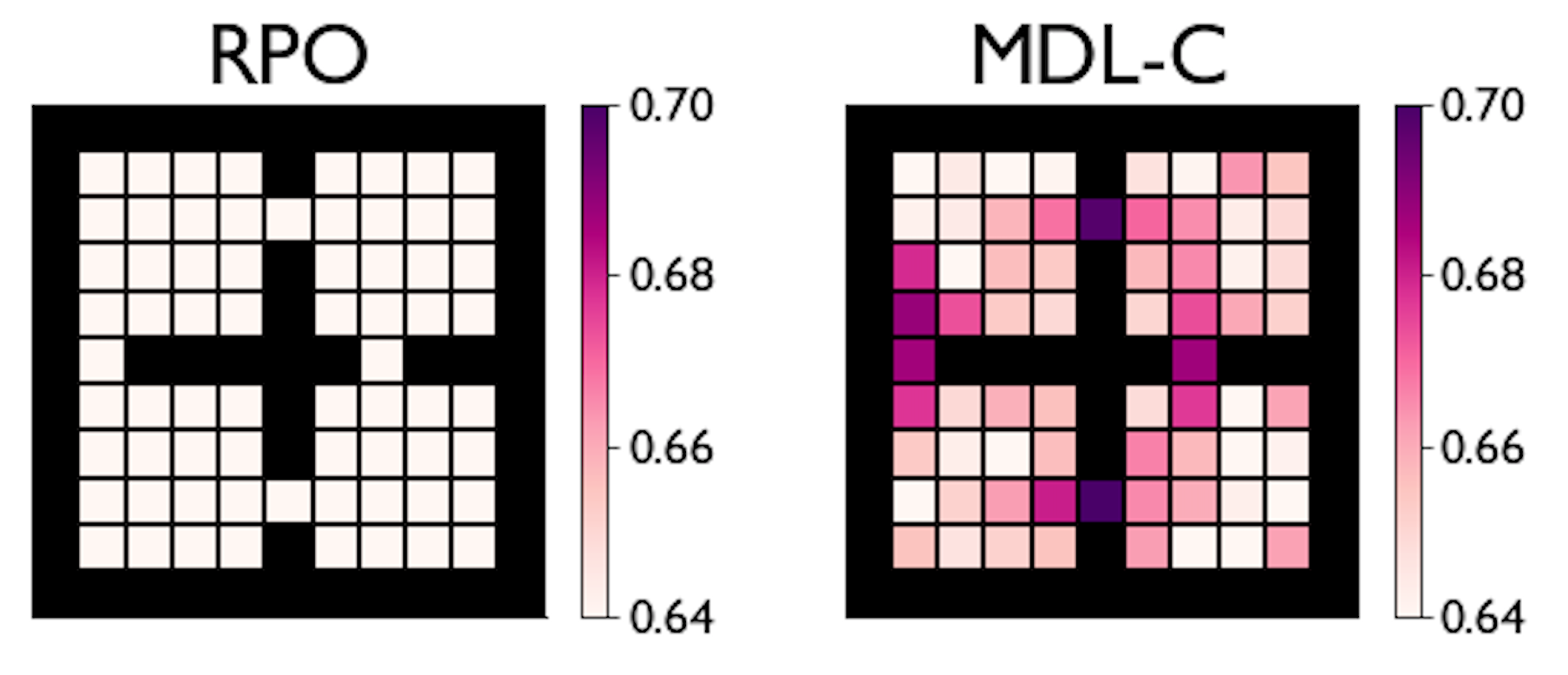}
  \caption{Heatmaps of  $\kl[\pi_\theta(\cdot|s), \pi_{w}(\cdot|s)]$ $\forall s\in\St$ for RPO and  $\kl[\pi_\theta(\cdot|s), \pi_{\bar w}(\cdot|s)]$ $\forall s\in\St$, where $\bar w = \E_\nu w$ for MDL-C, averaged over all possible goal states. The RPO default policy nearly perfectly matches the control policy, while the MDL-C default policy diverges most strongly from the control policy at the doorways. This is because the direction chosen by the policy in the doorways is highly goal-dependent. Because the MDL-C default policy learns to ignore the goal feature, it's roughly uniform in the doorways, whereas the control policy is highly deterministic, having access to the goal feature.}
  \label{fig:4rooms_kl}
\end{figure*}

\subsection{DeepMind Control Suite}

\begin{figure*}[ht]
  \centering
  \includegraphics[width=0.99\textwidth]{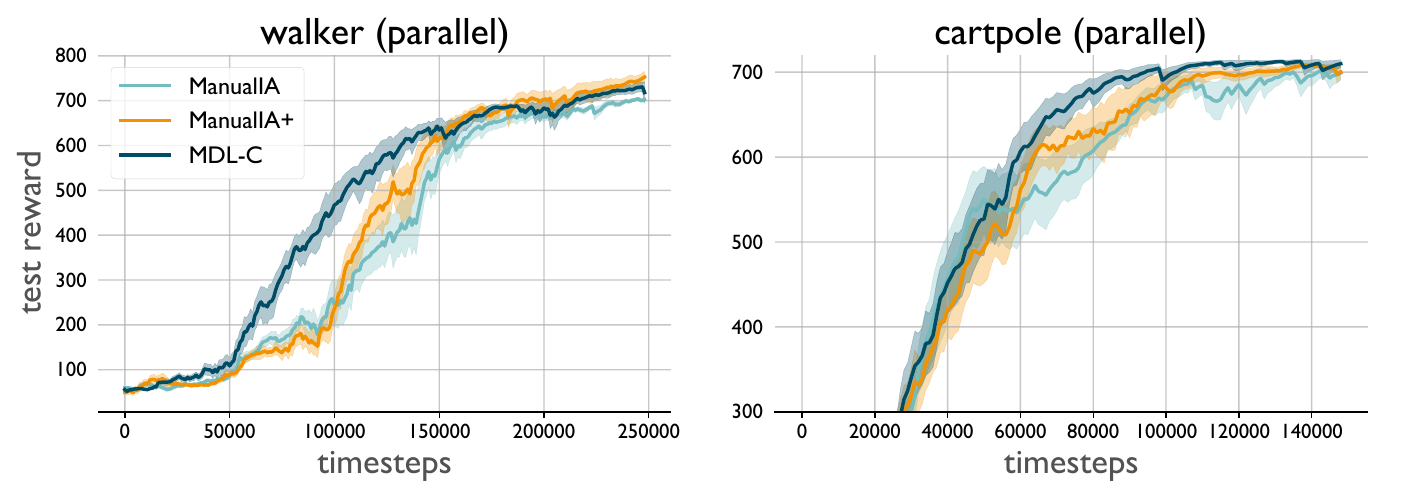}
  \caption{To test the effect of information asymmetry on its on performance, we trained a variant of \textsc{ManualIA} in which we withheld the input features that MDL-C learned to gate out (\cref{fig:dmc}) in addition to the task ID feature. We call this modified method \textsc{ManualIA+}. Average performance is plotted above over 10 seeds, with the shading representing one unit of standard error. We can see that while \textsc{ManualIA+} narrowly outperforms \textsc{ManualIA}, the performance gains of MDL-C can't solely be ascribed to effective information asymmetry.}
  \label{fig:manual-ia_plus}
\end{figure*}

\begin{figure*}[ht]
  \centering
  \includegraphics[width=0.99\textwidth]{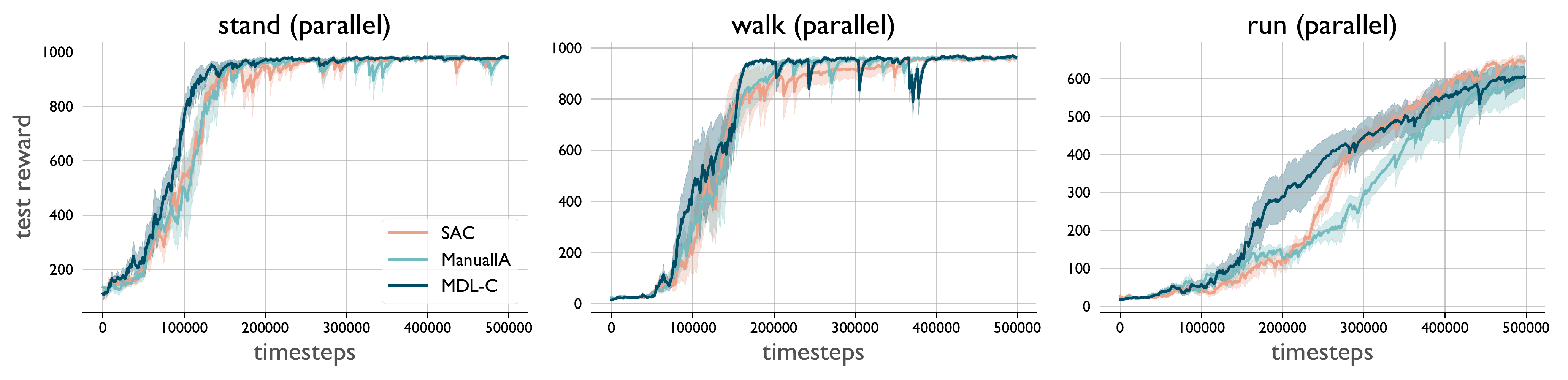}
  \caption{Test reward on each individual task in the \texttt{walker} domain over the course of parallel task training. Average performance is plotted above over 10 seeds, with the shading representing one unit of standard error. We can see the biggest performance difference on \texttt{walker, run}, the most challenging task.}
  \label{fig:walker_mt_individual}
\end{figure*}

\begin{figure*}[ht]
  \centering
  \includegraphics[width=0.99\textwidth]{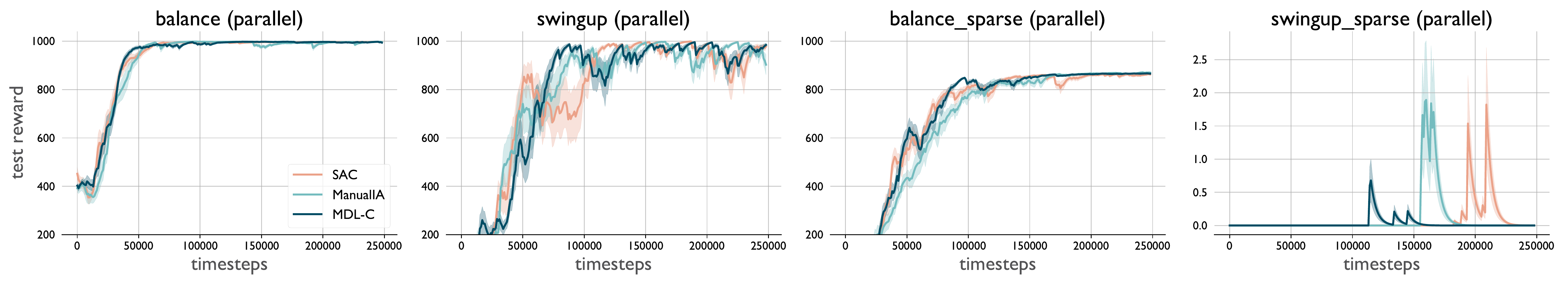}
  \caption{Test reward on each individual task in the \texttt{cartpole} domain over the course of parallel task training. Average performance is plotted above over 10 seeds, with the shading representing one unit of standard error. Interestingly, unlike in the sequential learning setting, joint training seems to impede performance on \texttt{swingup\_sparse}, with no method succeeding.}
  \label{fig:cartpole_mt_individual}
\end{figure*}

\end{appendices}

\clearpage

\bibliographystyleapp{apalike}
\bibliographyapp{appendix}

\end{document}